\documentclass[10pt,twocolumn,letterpaper]{article}

\usepackage{cvpr}              

\usepackage[dvipsnames]{xcolor}
\usepackage{multirow}

\definecolor{cvprblue}{rgb}{0.21,0.49,0.74}
\usepackage[pagebackref,breaklinks,colorlinks,citecolor=cvprblue]{hyperref}
\usepackage{graphicx}
\usepackage{wrapfig}
\usepackage{caption}
\usepackage[export]{adjustbox}
\usepackage{bm}
\usepackage{arydshln}
\usepackage{multirow}
\usepackage{amsmath}
\usepackage[export]{adjustbox}
\usepackage{bm}
\usepackage[accsupp]{axessibility} 

\makeatletter
\def\@fnsymbol#1{\ensuremath{\ifcase#1\or \dagger\or \ddagger\or
   \mathsection\or \mathparagraph\or \|\or **\or \dagger\dagger
   \or \ddagger\ddagger \else\@ctrerr\fi}}
\makeatother

\def\paperID{14071} 
\def\confName{CVPR}
\def\confYear{2024}

\title{Pre-trained Vision and Language Transformers Are \\ Few-Shot Incremental Learners}

\author{Keon-Hee Park$^{1}$
\and
Kyungwoo Song$^{2}$\thanks{Corresponding authors}
\and 
Gyeong-Moon Park$^{1}$\footnotemark[1]
\vspace{2mm} \and
$^{1}$Kyung Hee University, Republic of Korea \\
$^{2}$Yonsei University, Republic of Korea \\
{\tt\small \{pgh2874, gmpark\}@khu.ac.kr} \quad\quad
{\tt\small kyungwoo.song@yonsei.ac.kr}
}

\begin{document}
\maketitle
\begin{abstract}
Few-Shot Class Incremental Learning (FSCIL) is a task that requires a model to learn new classes incrementally without forgetting when only a few samples for each class are given. FSCIL encounters two significant challenges: catastrophic forgetting and overfitting, and these challenges have driven prior studies to primarily rely on shallow models, such as ResNet-18. Even though their limited capacity can mitigate both forgetting and overfitting issues, it leads to inadequate knowledge transfer during few-shot incremental sessions. In this paper, we argue that \textit{large models such as vision and language transformers pre-trained on large datasets can be excellent few-shot incremental learners.} To this end, we propose a novel FSCIL framework called \textbf{PriViLege}, \textbf{Pr}e-tra\textbf{i}ned \textbf{Vi}sion and \textbf{L}anguage transformers with prompting functions and knowl\textbf{e}d\textbf{ge} distillation. Our framework effectively addresses the challenges of catastrophic forgetting and overfitting in large models through new pre-trained knowledge tuning (PKT) and two losses: entropy-based divergence loss and semantic knowledge distillation loss.
Experimental results show that the proposed PriViLege significantly outperforms the existing state-of-the-art methods with a large margin, \textit{e.g.}, $+9.38\%$ in CUB200, $+20.58\%$ in CIFAR-100, and $+13.36\%$ in miniImageNet. 
Our implementation code is available at \url{https://github.com/KHU-AGI/PriViLege}.
\end{abstract}
\section{Introduction}
\label{sec:intro}
We humans have an exceptional ability to quickly comprehend novel concepts from only a small amount of data. To grant this ability for deep neural networks, Few-Shot Class Incremental Learning (FSCIL), introduced in~\cite{tao2020few} first, imitates a way of learning that closely resembles that of human learning. FSCIL typically comprises a base session and incremental sessions. During the base session, a network learns numerous classes with sufficient training data, while in the incremental sessions, it trains novel classes with few-shot training data per each class. Given the restricted amount of data in incremental sessions, an effective transfer of diverse knowledge learned in the base session is crucial in FSCIL.

In FSCIL, there are two significant challenges: catastrophic forgetting~\cite{mccloskey1989catastrophic} and overfitting~\cite{tao2020few}. Catastrophic forgetting occurs while the network learns new classes sequentially, \ie, the network severely forgets the previously learned knowledge. On the other hand, overfitting arises when the network overly focuses on a limited set of training data, resulting in a degradation of overall performance. To address these challenges, previous studies mainly have utilized shallow models like Resnet-18~\cite{kim2023warping,zhang2021few,yangneural}. The advantage of adopting a shallow model lies in its limited number of learnable parameters, making it effective for mitigating forgetting through partial freezing and curbing overfitting. However, the limited capacity of the shallow model hinders capturing and transferring sufficient domain knowledge from the base session to the incremental sessions.

Recently, large pre-trained models like Vision Transformer (ViT)~\cite{dosovitskiy2021an} and Contrastive Language-Image Pre-training (CLIP)~\cite{radford2021learning} are widely used in computer vision due to their promising adaptability and performance.
In that sense, large pre-trained models can effectively learn and transfer domain knowledge from the base session, overcoming limitations in transferability associated with shallow models.
However, finetuning large pre-trained models is prone to forget the useful pre-trained knowledge, while freezing the models hinders the acquisition of domain-specific knowledge during the base session. This inherent trade-off between preserving the pre-trained knowledge and acquiring new domain-specific knowledge hinders their use in FSCIL.

\begin{figure}[t]
    \centering
    \includegraphics[width=0.8\columnwidth]{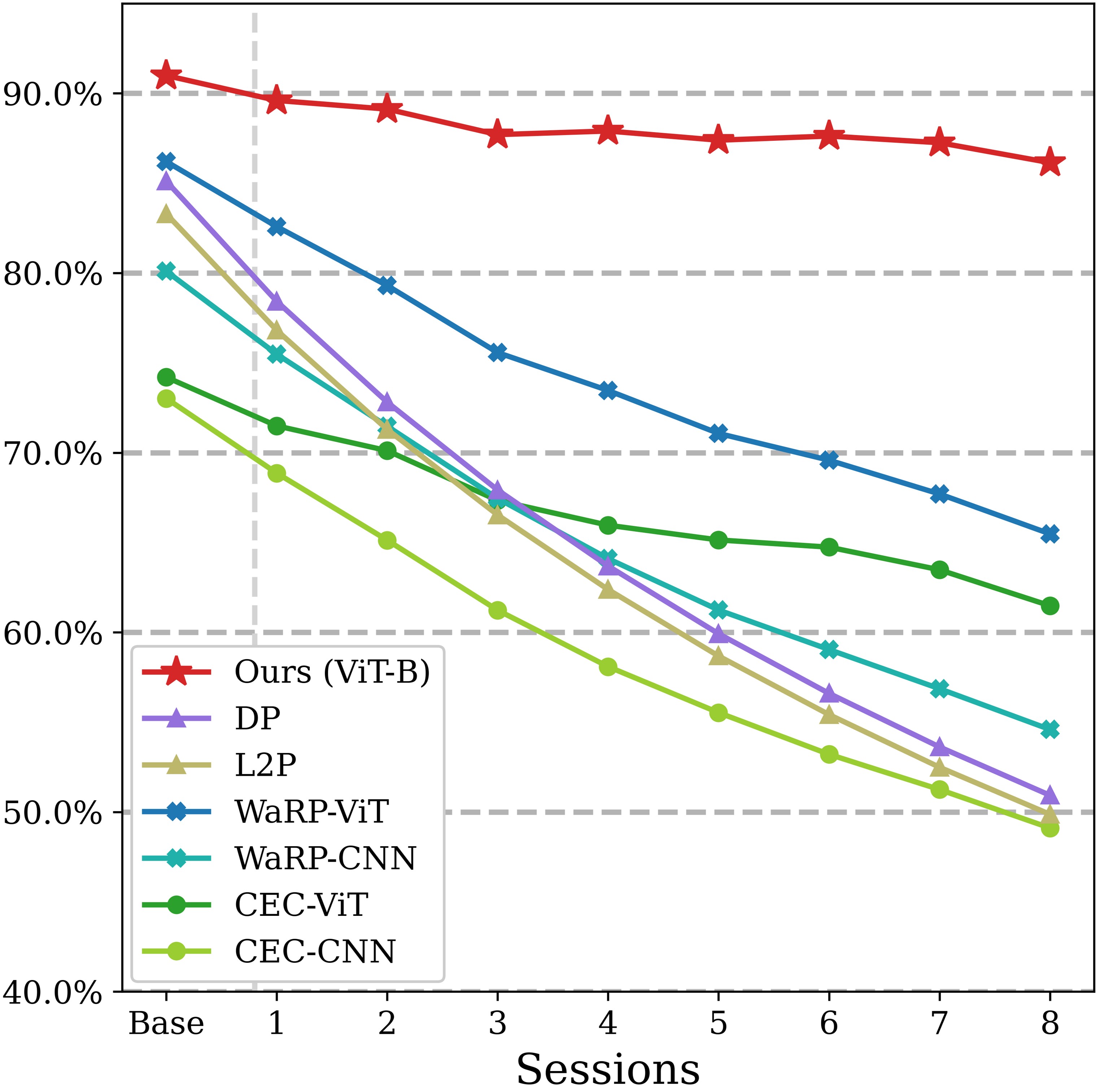}
    \vspace{-3mm}
    \caption{5-way 5-shot FSCIL experiments on CIFAR-100.}
    \vspace{-5mm}
    \label{fig:motiv}
\end{figure}

To investigate the challenges and applicability of large models in FSCIL, we conducted 5-way 5-shot experiments on CIFAR-100~\cite{krizhevsky2009learning}, using a Vision Transformer base (ViT-B) pre-trained on ImageNet-21K~\cite{ILSVRC15}. We applied this pre-trained model to existing FSCIL methods~\cite{kim2023warping,zhang2021few}. As shown in Figure~\ref{fig:motiv}, we find that directly applying the ViT backbone to existing methods is ineffective in FSCIL. In detail, selectively freezing parameters (WaRP~\cite{kim2023warping}) leads to severe forgetting during incremental sessions.
On the other hand, freezing the entire network (CEC~\cite{zhang2021few}) somewhat alleviates forgetting during the incremental sessions, but struggles to capture the useful knowledge in all the sessions. To sum up, existing FSCIL methods based on the large pre-trained model still show large performance drop due to catastrophic forgetting and the loss of transferability.

Recently, the methods utilizing prompt tuning~\cite{lester2021power} (L2P~\cite{wang2022learning}) or prefix tuning~\cite{li-liang-2021-prefix} (Dualprompt~\cite{wang2022dualprompt}) based on the pre-trained ViT show promising performances in Class Incremental Learning (CIL).
To further clarify the applicability of these methods in FSCIL, we conducted the experiments under the same setup in Figure~\ref{fig:motiv}. We observed that despite of effectively utilizing the large pre-trained model through prompting functions, these methods exhibit inferior performances compared to existing FSCIL methods. We attribute this to the limited number of learnable parameters in the prompt , which hinders effective knowledge transfer to incremental sessions. In other words,
the sole utilization of prompting functions to the pre-trained ViT is inadequate for transferring the sufficient domain knowledge in FSCIL.

In this paper, we argue that \textit{large models such as vision and language transformers pre-trained on large datasets can be excellent few-shot incremental learners}. To this end, we propose a new FSCIL framework based on \textbf{Pr}e-tra\textbf{i}ned \textbf{Vi}sion and \textbf{L}anguage transformers with prompting functions and knowl\textbf{e}d\textbf{ge} distillation, called \textbf{PriViLege}.
Our framework includes newly proposed Pre-trained Knowledge Tuning (PKT), which is a simple yet effective approach to preserve the pre-trained knowledge of large models while learning the domain-specific knowledge effectively.
Specifically, our PKT selectively trains specific layers with a new prompt modulation approach to prevent severe forgetting and enhance the knowledge acquisition of prompt.
To strengthen the discriminative representation learning during the base session, we introduce a novel entropy-based divergence loss. In addition, we propose a new knowledge distillation, utilizing the pre-trained language model (PLM) to transfer semantic knowledge from the language space to the visual space. Through extensive experiments, we demonstrate that our framework enables the pre-trained large models to effectively serve as few-shot incremental learners with significant improvement. \\ Our main contributions can be summarized as follows: 
\vspace{-0.1mm}
\begin{itemize}
 \setlength\itemsep{2mm}
  \item To address the challenges of adopting large pre-trained models in FSCIL, we propose a novel framework \textbf{PriViLege}, \textbf{Pr}e-tra\textbf{i}ned \textbf{Vi}sion and \textbf{L}anguage transformers with prompting functions and knowl\textbf{e}d\textbf{ge} distillation.
  \item We propose a new pre-trained knowledge tuning (PKT), which is designed to obtain domain knowledge effectively during the base session while preserving the useful pre-trained knowledge.

  \item To enhance the discriminative power during the base session and transfer the knowledge into the incremental sessions, we propose a new entropy-based divergence loss.
  \item We propose a semantic knowledge distillation loss to enhance representation learning by distilling semantic knowledge captured from the pre-trained language model.
  \item Comprehensive experiments show that our framework achieves overwhelming performance gains in the FSCIL benchmarks compared to state-of-the-art models.
  
\end{itemize}
\vspace{-2mm}

\section{Related Work}
\label{gen_inst}
\paragraph{Few-Shot Class Incremental Learning.}
Few-shot class incremental learning (FSCIL) is a sort of class incremental learning but is more challenging since a model is able to learn novel classes with only few training samples.
Among many previous approaches for FSCIL~\cite{tian2023survey},
dynamic network structure-based methods~\cite{tao2020few, zhang2021few, yan2021dynamically, seo2023lfs} 
adjust the network structures themselves during training, while preserving the severe forgetting of the previous knowledge. Feature and feature space-based methods~\cite{zhou2022forward,peng2022few,zhou2022few,shi2021overcoming,akyureksubspace,kim2023warping} enable the model to adapt to new classes better and improve the generalization ability of feature extractors for new classes. Prototype-based methods~\cite{mazumder2021few,hersche2022constrained,yangneural} aim to align prototypes with classifier weights to enhance the classification performance. However, prior methods primarily address forgetting and overfitting in shallow networks, leading to marginal performance improvements given the limited model capacity.
In this paper, we adopt large models such as the pre-trained ViT and CLIP for FSCIL and introduce how to utilize them effectively to overcome major challenges of FSCIL.

\vspace{-4mm}
\paragraph{Prompt Engineering for Vision Transformer.}
Prompt tuning~\cite{lester2021power} and prefix tuning~\cite{li-liang-2021-prefix} are widely used for prompt engineering in vision transformer. Prompt tuning adds learnable prompts to the input sequence, while prefix tuning directly influences attention patterns by appending prompts to the key and value for task-specific knowledge acquisition. Vision transformers utilizing prompt engineering show remarkable performances in class incremental learning. L2P~\cite{wang2022learning} and Dualprompt~\cite{wang2022dualprompt} utilize prompt and prefix tunings for each to learn new classes while freezing the pre-trained ViT. L2P and Dualprompt leverages randomly initialized prompts for prompt engineering.
Some recent approaches~\cite{Smith_2023_CVPR,tang2022learning,jung2023generating} have suggested methods for generating prompts to adapt the domain space for effective continual learning. CODA-Prompt~\cite{Smith_2023_CVPR} requires a collection of prompt components that are combined with input-dependent weights to generate input-specific prompts.
APG~\cite{tang2022learning} and DAP~\cite{jung2023generating} utilize prompt generators comprising multiple components, including cross-attention layers, groups of learnable parameters, and linear layers. These prompt-generating methods demand extra components and training costs for the prompt generator.
Unlike existing prompt-generating methods that necessitate additional learnable components for prompt generation, we propose a lightweight modulation approach for prompt. This approach significantly reduces the requirement of extra trainable components, and at the same time, it enhances the representation learning of prompt.

\vspace{-4mm}
\paragraph{Semantic Guidance from Language Models.}
The utilization of language embeddings for effective learning of novel classes has been extensively explored in the context of generalized zero-shot learning~\cite{akata2013label, pourpanah2022review}. Recently, a trial by \cite{khan2023introducing} investigates the use of language guidance for enhancing representation learning in continual learning. In the field of FSCIL, various approaches have incorporated additional language information, particularly class names, to improve representation learning in the base session. For instance, \cite{zhou2022few} addresses the drift of classifier weights by calibrating the encoded language information between existing and new classes. Furthermore, \cite{akyureksubspace} proposes a regularization method leveraging the relational information derived from class word embeddings extracted from the GloVe network~\cite{pennington2014glove}.
\begin{figure*}[t!] 
    \begin{center}
        \centerline{\includegraphics[width=\textwidth]{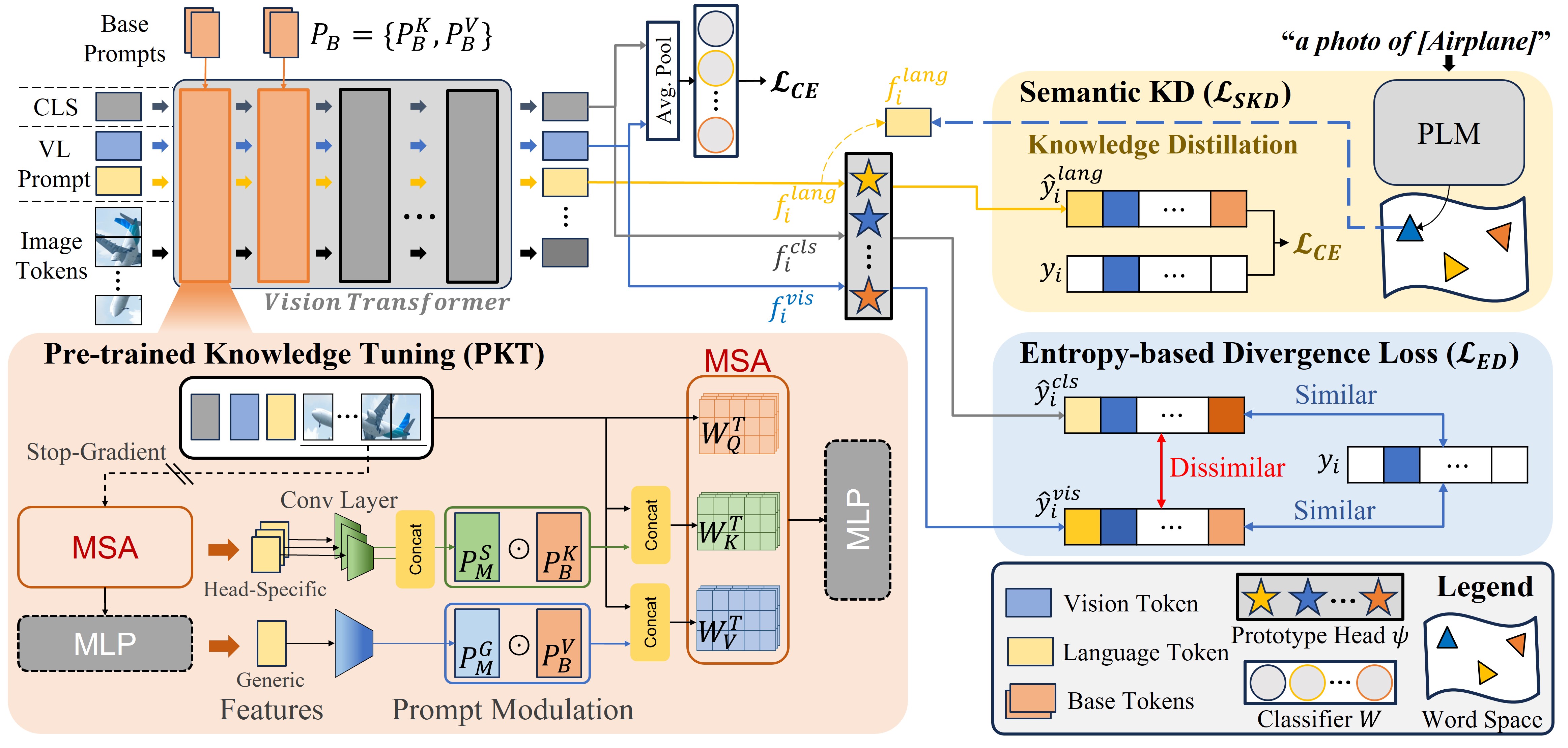}}
    \end{center}
    \vspace{-12mm}
    \caption{Overall framework of our method. In the base session, the newly proposed PKT trains the B-Prompt, VL-Prompt, and selected layers in the pre-trained ViT. $\mathcal{L}_{ED}$ drives the vision token in VL-Prompt to enhance discriminative ability for better classification. $\mathcal{L}_{SKD}$ leverages language embeddings to provide semantic knowledge to the language token in VL-Prompt.}
    \label{fig:main}
    \vspace{-3mm}
\end{figure*}

\section{Method}
\paragraph{Formulation of Few-Shot Class Incremental Learning.}
In FSCIL, the training dataset $\mathcal{D}=\{ \mathcal{D}^{0}, \mathcal{D}^{1}, ..., \mathcal{D}^T\}$ are sequentially given, where $\mathcal{D}^{0}$ is the dataset for the base session, $\mathcal{D}^t=\left\{ \left ( x_i,y_i \right )\right\}_{i=1}^{\left| \mathcal{D}^t\right|}$ is the dataset for the $t$-th incremental session, $1 \leq t \leq T$, and $T$ is total number of incremental sessions, respectively. Here, $\mathcal{D}^{0}$ for the base session generally has a large label space $\mathcal{C}^{0}$ and enough training data for each class $c \in \mathcal{C}^{0}$. On the other hand, $\mathcal{D}^t$ for the incremental session has only few training samples per each class, \textit{i.e.}, $\left| \mathcal{D}^{t} \right|={{k}\cdot\left| \mathcal{C}^t\right|}$, where ${\left| \mathcal{C}^t\right|}$ is total number of novel classes for the t-th task and $k$ is the number of samples per novel class. There is no class overlap between sessions, and at each session, the model can access the current dataset only. In this setting, the goal of FSCIL is to enable the model to incrementally learn new classes from a few samples, while preserving the classification ability of all previously encountered classes.

\vspace{-2mm}
\paragraph{Method Overview.}
Figure \ref{fig:main} provides an overview of the proposed method, where the pre-trained Vision Transformer (ViT) serves as the backbone network. To refine the pre-trained knowledge of ViT, we introduce a novel Pre-trained Knowledge Tuning (PKT) (Section \ref{sec:PKT}). PKT involves training a base prompt (B-Prompt), a vision-language prompt (VL-Prompt), and selected layers of ViT, thereby enhancing the transferable knowledge for incremental sessions. Additionally, to strengthen the discriminative ability during the base session, we propose an entropy-based divergence loss (Section \ref{sec:ED}) for a vision token in VL-Prompt. Finally, we introduce a semantic knowledge distillation loss (Section \ref{sec:SKD}) to transfer the semantic knowledge into a language token in VL-Prompt, improving the representation learning of our model. For stable learning in the few-shot environment, we utilize prototypes of each class as a classifier.
\subsection{Pre-trained Knowledge Tuning}
\label{sec:PKT}
Recently, the large pre-trained ViT based on prompting functions~\cite{wang2022learning, wang2022dualprompt, moon2023online} shows remarkable performance in class incremental learning. However, the effective integration of large pre-trained ViT into FSCIL remains unexplored. Existing methods adopting the pre-trained ViT struggle with issues such as catastrophic forgetting and overfitting. Furthermore, previous prompt-based methods face challenges in transferring sufficient knowledge to the incremental sessions due to the limited prompt size. 

In this paper, we explore how to adapt the powerful pre-trained ViT to the FSCIL task effectively.
To this end, we introduce a novel approach termed Pre-trained Knowledge Tuning (PKT), which selectively fine-tunes specific layers using additional prompts to acquire domain-specific knowledge during the base session. In PKT, we specifically update the initial $L$ layers of the pre-trained ViT $f_\theta$, where $L$ is a hyperparameter representing the number of layers to be updated. Through empirical analysis, we determined the optimal number of layers to be updated, \textit{e.g.}, the first two layers $(L=2)$ in ViT-B. Since we freeze most layers in ViT, the pre-trained knowledge remains helping incremental learning without forgetting. The updated ViT is frozen after the base session to preserve domain-specific knowledge and transfers the learned knowledge to incremental sessions.
Since we freeze most of the layers, 
effective learning of domain-specific knowledge becomes a challenge.
To address this limitation,
we introduce two key prompts: the base prompt (B-Prompt) denoted as $ \bm{P_{B}}\in \mathbb{R}^{L \times 2 \times D}$, where $D$ represents the embedding dimension, and the vision-language Prompt (VL-Prompt) denoted as $\bm{P_{VL}} \in \mathbb{R}^{2 \times D}$. B-Prompt is tailored to capture domain-specific knowledge while selectively fine-tuning some layers at the base session. B-Prompt facilitates the transfer of domain-specific knowledge to incremental sessions. Meanwhile, VL-Prompt, consisting of both vision and language tokens, is designed to transfer positive knowledge from all previous sessions to the next. We train B-Prompt and VL-Prompt using the prefix tuning and the prompt tuning, respectively. Through the utilization of both prefix tuning and prompt tuning, we are able to tailor the training of B-Prompt and VL-Prompt to their respective purposes.

However, since prefix tuning has a limitation of updating B-Prompt as mentioned in~\cite{gao2023lae,he2022towards} which is the slow adaptation speed of learnable prompts, B-Prompt cannot properly learn domain-specific knowledge and can be ignored by the fine-tuned layers. To overcome this, we propose new modulation prompts $P_M$ to assist B-Prompt. Modulation prompts contain a head-specific prompt $P_M^S$ and a generic prompt $P_M^G$, which are obtained from the multi-head self-attention (MSA) layer and the followed MLP layer of the pre-trained ViT, respectively. The formulation of modulation prompts is as follows:
\begin{align}
    h^{MSA}&= \mathrm{MSA}(h_Q,h_K,h_V), \\[2pt]
    P_M^S&=[g^S_{1}(h^{MSA}_{1});...\,;g^S_{H}(h^{MSA}_{H})], \\
    h^{MLP}&= \mathrm{MLP}(h^{MSA}), \\
    P_M^G&= g^G(h^{MLP}),
\end{align}
where $H$ denotes the number of heads in the MSA layer, $h^{MSA}$ and $h^{MLP}$ denote the outputs for each layers, $g^S=\{g^S_1,\cdots,g^S_H\}$ and $g^G$ denote point-wise convolution layers for $P_M^S$ and $P_M^G$, respectively. We extract the feature vectors from the ViT layers and generate modulation prompts through $1\times1$ convolution layers to align with the B-Prompt.

Through the pre-trained layers and input data, both the head-specific prompt and generic prompt enhance feature knowledge. These modulation prompts can scale the B-Prompt, enlarging its feature vector depending on the input data. The modulation prompts assist the B-Prompt in capturing domain-specific knowledge by enlarging feature vectors.
The process of prefix tuning of PKT is as follows:
\begin{align}
    &\bar P_K^{'} = P_M^S \odot P_B^K,\\
    &\bar P_V^{'} = P_M^G \odot P_B^V,\\
    &\bar h^{out}=\mathrm{MSA}([P_{VL}^Q;h_Q],[\bar P_K^{'};h_K],[\bar P_V^{'};h_V]),
\end{align}
where $\bar h^{out}$ denotes the output of the PKT, and $h_Q$,$h_K$, and $h_V$ denote the input query, key, and value, respectively.

In summary, our PKT provides two main advantages: 1) it effectively learns base session knowledge by introducing additional plasticity in the first $L$ layers and incorporating extra prompts, and 2) by scaling the B-Prompt through the modulation prompts, PKT promotes the update of the B-Prompt. This boosts the B-Prompt to learn useful domain-specific knowledge along with the pre-trained ViT. Empirical results demonstrate that PKT significantly enhances performance, facilitating the positive knowledge transfer in incremental sessions.

\subsection{Entropy-based Divergence Loss}
\label{sec:ED}
During the training of prompts and selected layers using PKT, our model effectively acquires the domain knowledge, for its positive transfer during incremental sessions. To embed multi-perspective knowledge, we involve the vision token along with the [CLS] token for the classification through average pooling.
However, since these two tokens share the same objective, their output features become similar as training progresses, which hinders the effective training of the vision token.
To strengthen the discriminative power of the vision token itself, we propose a new regularization term, called an entropy-based divergence loss~($\mathcal{L}_{ED}$).

To calculate $\mathcal{L}_{ED}$, we first construct a prototype classifier $\psi \in \mathbb{R}^{|C^0|\times D}$ that consists of prototypes for the base session classes.
For each base class $c_j$, where $c_j \in C^0$, and $j \in \{1, 2, ..., |C^0|\}$, the prototype $proto_{c_j} \in \mathbb{R}^{D}$ is the average vector of all the output features extracted by the [CLS] token passing through the pre-trained ViT. Therefore, the prototype classifier $\psi$ can be represented as follows:
\begin{align}
     & proto_{c_j} = \frac{1}{N_{c_j}}\sum_{k=1}^{N_{c_j}}{f_k^{cls}},    \\
     & \mathbf{\psi} =[proto_{c_1};proto_{c_2};...\,;proto_{c_{|C^0|}}],
\end{align}
where $N_{c_j}$ is the number of training samples for the class $c_j$. Using the prototype classifier $\psi$, we then calculate the logits $\hat{y}^{cls}_i = \psi(f_i^{cls})$ and $\hat{y}_i^{vis} = \psi(f_i^{vis})$ corresponding to the [CLS] and vision tokens, respectively. At this time, the prototype classifier $\psi$ is not trainable to serve the stable basis to calculate the loss function. Finally, we use $\hat{y}^{cls}_i$ and $\hat{y}^{vis}_i$ with the label $y_i$ to define the entropy-based divergence loss $\mathcal{L}_{ED}$ as follows:
\begin{align}
     & \mathcal{L}_{ED} = \log(\frac{\mathcal{L}_{CE}(\hat{y_i}^{vis},y_i)+\mathcal{L}_{CE}(\hat{y_i}^{cls},y_i)}{\mathcal{L}_{KL}(\delta(\hat{y}_i^{vis}), \delta(\hat{y}_i^{cls}))} +1),
\end{align}
where $\mathcal{L}_{CE}$ is the cross entropy loss, $\mathcal{L}_{KL}$ is the Kullback-Leibler divergence loss~\cite{kullback1951information}, and $\delta (\cdot)$ is a softmax function. To minimize $\mathcal{L}_{ED}$, our model learns to minimize both $\mathcal{L}_{CE}(\hat{y_i}^{vis},y_i)$ and $\mathcal{L}_{CE}(\hat{y_i}^{cls},y_i)$, and maximize the $\mathcal{L}_{KL}(\delta(\hat{y}_i^{vis}), \delta(\hat{y}_i^{cls}))$.
In other words, the proposed entropy-based divergence loss guides the vision token to gain the discriminative knowledge itself, while separating the embedded knowledge of the vision token from the one of the [CLS] token. Through $\mathcal{L}_{ED}$, our model can capture the domain-specific knowledge effectively using the vision token at the base session and provide this transferable knowledge for the incremental sessions.

\begin{table*}[t]
    \resizebox{\textwidth}{!}{%
        \setlength{\tabcolsep}{1.5pt}
        \renewcommand{\arraystretch}{1.6}
        \begin{tabular}{ccccccccccccc}
            \specialrule{1.3pt}{1pt}{1pt}
            {\large Dataset}              &  & \multicolumn{3}{c}{\large CUB200} &                        & \multicolumn{3}{c}{\large CIFAR-100} &  & \multicolumn{3}{c}{\large miniImageNet}                                                                                                                                 \\ \cline{1-1} \cline{3-5} \cline{7-9} \cline{11-13}
            {\large Method}               &  & {\large $A_{Base}$}                  & {\large $A_{Last}$}    & {\large $A_{Avg}$}                      &  & {\large $A_{Base}$}               & {\large $A_{Last}$}    & {\large $A_{Avg}$}     &  & {\large $A_{Base}$}    & {\large $A_{Last}$}    & {\large $A_{Avg}$}     \\ \cline{1-1} \cline{3-5} \cline{7-9} \cline{11-13}
            Fine-Tuning + Proto $\psi$             &  & \textbf{84.21±0.13}                  & 3.79±1.47             & 21.60±1.32                              &  & \textbf{91.36±0.15}            & 5.19±0.13             & 37.04±1.06             &  & \underline{93.67±0.02}             & 9.87±5.42              & 44.60±0.92             \\ \cdashline{1-1} \cdashline{3-5} \cdashline{7-9} \cdashline{11-13}
            CEC [CVPR’21]                 &  & 75.40±8.01                          & \underline{65.70±8.03}             & \underline{72.41±1.18}                              &  & 74.20±2.03                        & 61.48±3.33 & 67.10±2.92 &  & 87.43±5.90             & \underline{80.74±7.51} & \underline{83.06±7.14} \\
            L2P [CVPR’22]                 &  & 44.97±2.32                           & 15.41±3.45             & 24.99±4.30                              &  & 83.29±0.50                        & 49.87±0.31             & 64.08±0.39             &  & 94.59±0.21             & 56.84±0.32             & 72.97±0.36             \\
            DualPrompt [ECCV’22]          &  & 53.37±1.83                           & 23.25±2.02             & 36.30±2.39                              &  & 85.11±0.29                        & 50.93±0.21             & 65.45±0.27             &  & 95.05±0.20             & 57.14±0.11             & 73.31±0.15             \\
            NC-FSCIL [ICLR’23]                &  & 78.49±2.32                           & 38.80±1.14 & 57.92±1.71                              &  & 89.51±0.23                        & 53.70±0.14             & 68.96±0.17             &  & 77.25±0.42             & 46.35±0.25             & 59.52±0.33             \\
            WaRP [ICLR’23]            &  & 67.74±5.57                           & 49.36±6.56             & 55.85±6.06                  &  & 86.20±1.46                        & \underline{65.48±1.87}             & \underline{74.55±1.67}             &  & 83.30±1.06 & 67.97±1.28             & 74.13±1.08             \\ \cdashline{1-1} \cdashline{3-5} \cdashline{7-9} \cdashline{11-13}
            \textbf{PriViLege (Ours)} &  & \underline{82.21±0.35}                           & \textbf{75.08±0.52}    & \textbf{77.50±0.33}                     &  & \underline{90.88±0.20}                        & \textbf{86.06±0.32}    & \textbf{88.08±0.20}    &  & \textbf{96.68±0.06}    & \textbf{94.10±0.13}    & \textbf{95.27±0.11}    \\
            \specialrule{1.3pt}{1pt}{1pt}
        \end{tabular}%
    }
    \vspace{-3mm}
    \caption{Comparison of the performance on CUB200, CIFAR-100, and miniImageNet. CUB200 has a 10-way 5-shot incremental setup, and CIFAR-100 and miniImageNet have a 5-way 5-shot incremental setup. We report the best as \textbf{bold} and the second-best as \underline{underlined}.\vspace{-3mm}}
    
    \label{tab:main}
    \vspace{-2mm}
\end{table*}

\subsection{Semantic Knowledge Distillation Loss}
\label{sec:SKD}
Even though the transferred knowledge from the base session to the incremental sessions is abundant and useful, it is still challenging to learn the exact representations for novel classes from few-shot training samples. To alleviate this issue,
it is required to provide external knowledge related to the novel classes for better adaptation. To this end, we introduce a new semantic knowledge distillation loss ($\mathcal{L}_{SKD}$) to provide additional semantic knowledge by using the pre-trained language model (PLM), e.g., BERT~\cite{kenton2019bert}. Through the language embeddings from PLM utilizing the class names given as labels, we can provide useful semantic knowledge to our proposed model.


To achieve the goal of semantic knowledge distillation, we first get the language embedding feature $w_{cn_i} = f_{\varphi}(word_{cn_i})$, where $f_{\varphi}$ is PLM and $word_{cn_i}$ is the class name prompted as ``\textit{a photo of} [$cn_i$]", corresponding to the class $cn_i$. Meanwhile, we also acquire the output feature $f_i^{lang}=f_\theta(x_i)[2]$ corresponding to the language token in VL-Prompt using the ViT backbone.
To distill the semantic knowledge from the language embedding feature $w_{cn_i}$ to the language token feature $f_i^{lang}$, we adopt the knowledge distillation loss ($\mathcal{L}_{KD}$) from~\cite{hinton2015distilling}. However, these two features come from totally different embedding spaces, \textit{i.e.}, visual feature space, and language embedding space, respectively, solely applying the distillation loss to match two different distributions may be ineffective.

To overcome this issue, we utilize the prototype classifier $\psi$ once again as a stable basis to regulate the output feature of the language token. Specifically, we input the language token feature $f_i^{lang}$ into the prototype classifier $\psi$, denoted as $\hat{y}_i^{lang}=\psi(f_i^{lang})$, to compute the cross-entropy loss. We utilize the cross-entropy loss ($\mathcal{L}_{CE}$) to minimize the distribution difference between the visual feature space and the language embedding space.
We then define semantic knowledge distillation loss $\mathcal{L}_{SKD}$ by adding two losses as follows:
\begin{align}
     \mathcal{L}_{SKD}=\mathcal{L}_{KD}(f_i^{lang},w_{cn_i}) + \gamma\cdot \mathcal{L}_{CE}(\hat{y}_i^{lang},y_i),
\end{align}
where $\gamma$ is the balancing hyperparameter for $\mathcal{L}_{CE}$. The second term in $\mathcal{L}_{SKD}$ prevents $f_i^{lang}$ from diverging to the undesirable feature representation using the true label $y_i$. We used $\gamma=0.1$ for all of our experiments.

To sum up, the proposed semantic knowledge distillation loss $\mathcal{L}_{SKD}$ enables our model to distill the useful semantic knowledge from the language embedding space into the visual feature space to provide additional information during the few-shot incremental sessions.
It is beneficial to mitigate the challenge of representation learning from the few-shot data. Moreover, $\mathcal{L}_{SKD}$ drives the network to learn abundant base knowledge using enough classes in the base session, which leads to positive knowledge transfer for the incremental sessions.

The total loss for the base session ($\mathcal{L}_{base}$) can be summerized as follows:
\begin{align}
     & \mathcal{L}_{base}=\mathcal{L}_{CE}(\hat{y}_i,y_i)+\alpha\cdot\mathcal{L}_{ED}+\beta\cdot\mathcal{L}_{SKD},
\end{align}
where $\alpha$ and $\beta$ are the scaling factors for entropy-based divergence loss and semantic knowledge distillation loss. In the incremental sessions, we do not use the entropy-based divergence loss since a few samples are not enough to learn discriminative features. Therefore, the total loss for the incremental sessions ($L_{inc}$) can be expressed as follows:
\begin{align}
     & \mathcal{L}_{inc}=\mathcal{L}_{CE}(\hat{y}_i,y_i)+\beta\cdot\mathcal{L}_{SKD}.
\end{align}

\section{Experiments}
\subsection{Experimental Settings}
\vspace{-2mm}
\paragraph{Datasets and Metrics.}
We evaluated our method with the SOTA FSCIL methods on three datasets: CIFAR-100~\cite{krizhevsky2009learning}, miniImageNet~\cite{ravi2017optimization}, and CUB200~\cite{wah2011caltech}.
As shown in Table~\ref{tab:exp-config}, we followed the same split configuration proposed by~\cite{tao2020few} in all datasets. We evaluated the performance by measuring the accuracy of the base session $A_{Base}$, last session $A_{Last}$, and the average accuracy of all the sessions $A_{Avg}$. We conducted 5 simulations under different random seeds and reported the averages.

\vspace{-4mm}
\paragraph{Baselines and Implementation Details.}
We considered the following recent FSCIL methods as baselines: CEC~\cite{zhang2021few}, WaRP~\cite{kim2023warping}, and NC-FSCIL~\cite{yangneural}. We also set L2P~\cite{wang2022learning} and DualPrompt~\cite{wang2022dualprompt} as baselines to compare the methods using ViT. For the backbone network, we used a ViT-B/16~\cite{dosovitskiy2021an} pre-trained by ImageNet-21K~\cite{ILSVRC15} for all the methods including ours. We used BERT-base~\cite{kenton2019bert} to extract the word class embeddings. We set the first two layers as trainable for the pre-trained knowledge tuning. We set 0.5 for both $\alpha$ and $\beta$. We used Adam optimizer~\cite{kingma2014adam}, cosine annealing scheduler~\cite{loshchilovsgdr}, and the learning rate as 2e-4. We trained our method using an RTX 3090 GPU and set the batch size as 128. We trained 5 epochs for the base session and 3 epochs for the incremental sessions.

\begin{table}[t]

\resizebox{\columnwidth}{!}{%
\setlength{\tabcolsep}{1.5pt}
\renewcommand{\arraystretch}{1.2}
\begin{tabular}{ccccccc}
\specialrule{1.3pt}{1pt}{1pt}
Session     &  & CUB200        &  & CIFAR-100        &  & miniImageNet     \\ \cline{1-1} \cline{3-3} \cline{5-5} \cline{7-7} 
Base        &  & 100            &  & 60               &  & 60               \\
Incremental &  & 10-way 5-shot &  & 5-way 5-shot &  & 5-way 5-shot \\  \cdashline{1-1} \cdashline{3-3} \cdashline{5-5} \cdashline{7-7}
\# of sessions       &  & 1+10           &  & 1+8              &  & 1+8              \\ 
\specialrule{1.3pt}{1pt}{1pt}
\end{tabular}%
}
\vspace{-3mm}
\caption{Configuration settings for FSCIL benchmarks on CUB-200, CIFAR-100, and miniImageNet. \vspace{-3mm}}
\label{tab:exp-config}
\vspace{-4mm}
\end{table}

\subsection{Main Experimental Results}
\vspace{-2mm}
We reported the base, last, and average accuracy of CUB200, CIFAR-100, and miniImageNet, respectively, in Table~\ref{tab:main}.
As shown in Table~\ref{tab:main}, our method, PriViLege, surprisingly overwhelmed all the baselines with a large margin on all the datasets compared with SOTA methods in FSCIL. Our proposed methods using ViT-B/16 reported an about $+9.38\%$ performance enhancement in $A_{Last}$ and about $+5.09\%$ in $A_{Avg}$ against CEC on CUB200. Our method also showed outstanding performance in CIFAR-100 where our proposed methods reported about $+20.58\%$ in $A_{Last}$ and about $+13.53\%$ in $A_{Avg}$ against WaRP. Also, compared with prompt-based methods such as L2P and DualPrompt, our novel method, PriViLege, reported powerful performance enhancement. Our experiments consistently revealed notable enhancements in $A_{Last}$ and $A_{Base}$ across all datasets. The emphasis on effective domain knowledge learning and transferability enhancement through PriViLege, incorporating PKT, $\mathcal{L}_{ED}$, and $\mathcal{L}_{SKD}$, contributes to its outstanding performance.

It is noteworthy that our method, PriViLege, showed significant performance improvement on CUB200. Given the fewer samples per class in CUB200, learning sufficient knowledge in the base session and transferring it to incremental sessions becomes challenging. However, our proposed method demonstrated improved performance across all metrics on CUB200. This underscores that the ability of PriViLege to capture effective domain-specific knowledge and transfer the knowledge into incremental sessions helps the successful application of ViT in FSCIL.

\begin{table}[t]
\resizebox{\columnwidth}{!}{%
\setlength{\tabcolsep}{5.pt}
\renewcommand{\arraystretch}{1.2}
\begin{tabular}{ccccccc}
\specialrule{1.3pt}{1pt}{1pt}
{\large Dataset}        &  & \multicolumn{5}{c}{{\large CUB 200}}                                                                                     \\ \cline{1-1} \cline{3-7} 
Ablation       &  & $A_{Base}$                           &  & $A_{Last}$                           &  & $A_{Avg}$                            \\ \cline{1-1} \cline{3-3} \cline{5-5} \cline{7-7} 
Baseline       &  & \textbf{84.21±0.13} &  & 3.79±1.47 &  & 21.60±1.32 \\ \cdashline{1-1} \cdashline{3-3} \cdashline{5-5} \cdashline{7-7} 
PKT            &  & 79.06±0.77 &  & 70.81±0.76 &  & 73.36±0.77 \\
PKT + $\mathcal{L}_{ED}$       &  & 80.31±0.54 &  & 72.70±0.45 &  & 75.04±0.40 \\
PKT + $\mathcal{L}_{SKD}$      &  & 82.10±0.57 &  & 73.44±0.40 &  & 76.27±0.30 \\ \cdashline{1-1} \cdashline{3-3} \cdashline{5-5} \cdashline{7-7} 
\textbf{Ours} &  & 82.21±0.35 &  & \textbf{75.08±0.52} &  & \textbf{77.50±0.33} \\ 
\specialrule{1.3pt}{1pt}{1pt}
\end{tabular}%
}
\vspace{-3mm}
\caption{Ablation experiment on CUB 200. The baseline denotes fine-tuning pre-trained ViT with prototype classifier $\psi$.}
\label{tab:ablation}
\vspace{-4mm}
\end{table}

\vspace{-2mm}
\subsection{Ablation Study}
\vspace{-2mm}
We conducted an ablation study on CUB200 to validate our method. We set fine-tuning pre-trained ViT with a prototype classifier $\psi$ as the baseline. Table~\ref{tab:ablation} illustrates the performance comparison of each component. PKT exhibited notable enhancements in $A_{Last}$ and $A_{Avg}$, showcasing the effectiveness of our proposed tuning approach in transferring domain-specific knowledge to incremental sessions. Despite limited fine-tuning layers by freezing most layers, our PKT showed only a slight decline in $A_{Base}$ compared to the baseline. This underscores that PKT can effectively capture domain-specific knowledge. Furthermore, our proposed losses, entropy-based divergence loss, and semantic knowledge distillation loss, demonstrated promising performance. Applying the entropy-based divergence loss resulted in a performance enhancement of approximately $+1.89\%$ in $A_{Last}$ and $+1.68\%$ in $A_{Avg}$. Similarly, the semantic knowledge distillation loss recorded improvements of about $+2.63\%$ in $A_{Last}$ and approximately $+2.91\%$ in $A_{Avg}$ when compared to the sole application of PKT. In conclusion, our method, PriViLege, demonstrated outstanding performance with significant margins compared to the baselines.

\begin{table}[t]
\resizebox{\columnwidth}{!}{%
\setlength{\tabcolsep}{5.pt}
\renewcommand{\arraystretch}{1.2}
\begin{tabular}{ccccccc}
\specialrule{1.3pt}{1pt}{1pt}
{Dataset}          &  & \multicolumn{5}{c}{CUB 200}                \\ \cline{1-1} \cline{3-7} 
\# of Layers &  & $A_{Base}$    &  & $A_{Last}$    &  & $A_{Avg}$     \\ \cline{1-1} \cline{3-3} \cline{5-5} \cline{7-7} 
0 Layers         &  & 76.07±0.56 &  & 60.19±1.11 &  & 67.08±0.71 \\ 
\textbf{2 Layers}         &  & \textbf{79.06±0.77} &  & \textbf{70.81±0.76} &  & \textbf{73.36±0.77} \\
5 Layers         &  & 78.42±0.84 &  & 68.52±0.84 &  & 71.99±0.84 \\
7 Layers         &  & 76.96±0.74 &  & 63.15±2.73 &  & 68.06±1.54 \\
10 Layers        &  & 74.95±0.78 &  & 57.78±2.30 &  & 64.19±1.59 \\
12 Layers        &  & 73.62±2.72 &  & 56.02±1.47 &  & 62.71±2.14 \\
\specialrule{1.3pt}{1pt}{1pt}
\end{tabular}%
}
\vspace{-3mm}
\caption{Further study on the number of tuned layers on CUB200.\vspace{-3mm}}
\vspace{-4mm}
\label{tab:Layertuning}
\end{table}
\vspace{-7mm}
\begin{figure}
    \centering
    \includegraphics[width=\columnwidth]{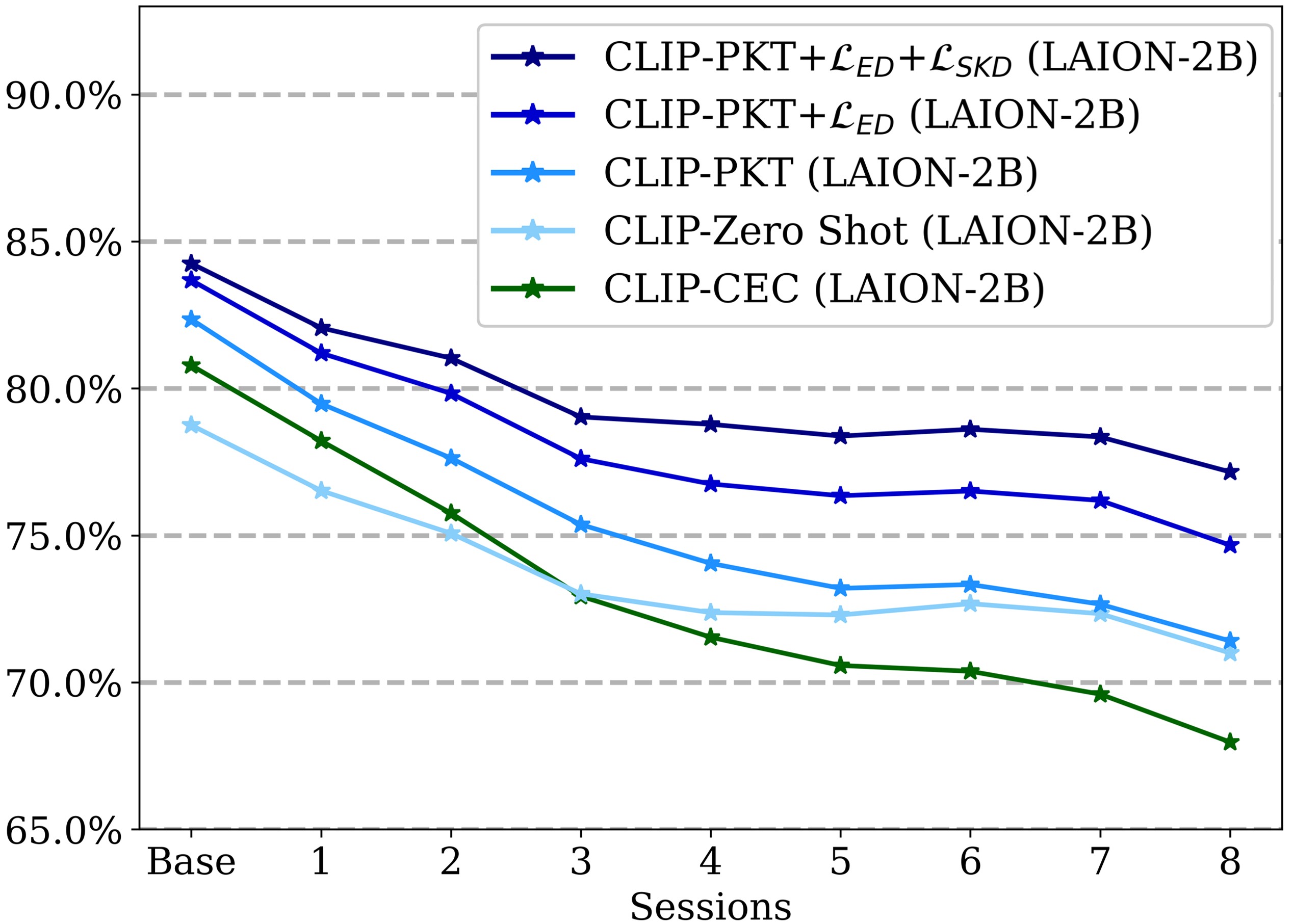}
    \vspace{-6mm}
    \caption{CLIP performance on CIFAR-100. We compare our proposed methods with zero-shot performance of CLIP and CEC.}
    \label{fig:analysis-clip}
\vspace{-6mm}
\end{figure}
\subsection{Analysis}
\vspace{-1mm}
\paragraph{PriViLege on Pre-trained CLIP Network.}
To evaluate the adaptability of our proposed method, PriViLege, we compared our method with the zero-shot performance of CLIP~\cite{radford2021learning}. To integrate PriViLege with CLIP, we exclusively trained the vision encoder of CLIP and employed a text encoder to extract language embedding features for the semantic knowledge distillation loss. Additionally, we compared the performance of the existing method, CEC, with the zero-shot performance in CLIP.

As shown in Figure~\ref{fig:analysis-clip}, We noticed that while the existing method, CEC, exhibited inferior performance compared to zero-shot performance, our novel method demonstrated outstanding results in contrast to zero-shot performance. Specifically, we observed that the proposed entropy-based divergence loss and semantic knowledge distillation loss significantly contributed to notable performance enhancements. Given that CLIP is pre-trained with a contrastive approach involving both vision and language data, our proposed entropy-based divergence loss can contribute to improved representation knowledge by enhancing discriminative knowledge. Moreover, since CLIP already incorporates language embedding features, the semantic knowledge distillation loss effectively provides external knowledge for better adaptation. Our experiments revealed that our proposed method is applicable to CLIP and can show outstanding performance within this framework.
\vspace{-3mm}
\paragraph{Layer Tuning Ablation for PKT.}
We studied further experiment on CUB200 to find out how many layers to be tuned for the proposed PKT. As shown in Table~\ref{tab:Layertuning}, fine-tuning the first 2 layers with additional prompts in PKT showed best performance across all metrics. This approach effectively captures domain-specific knowledge while preserving pre-trained knowledge. It is noteworthy that the absence of layer tuning resulted in inferior performance in $A_{Base}$ and $A_{Last}$ due to the limited capacity to learn domain-specific knowledge. Moreover, an increase in the number of learnable layers led to a decline in both $A_{Base}$ and $A_{Last}$. It demonstrate that when the number of learnable layers is sufficient for capturing domain-specific knowledge, additional randomly initialized prompts may impede representation learning.
\begin{figure}[t]
    \begin{subfigure}[b]{0.49\columnwidth}
    \includegraphics[width=\columnwidth]{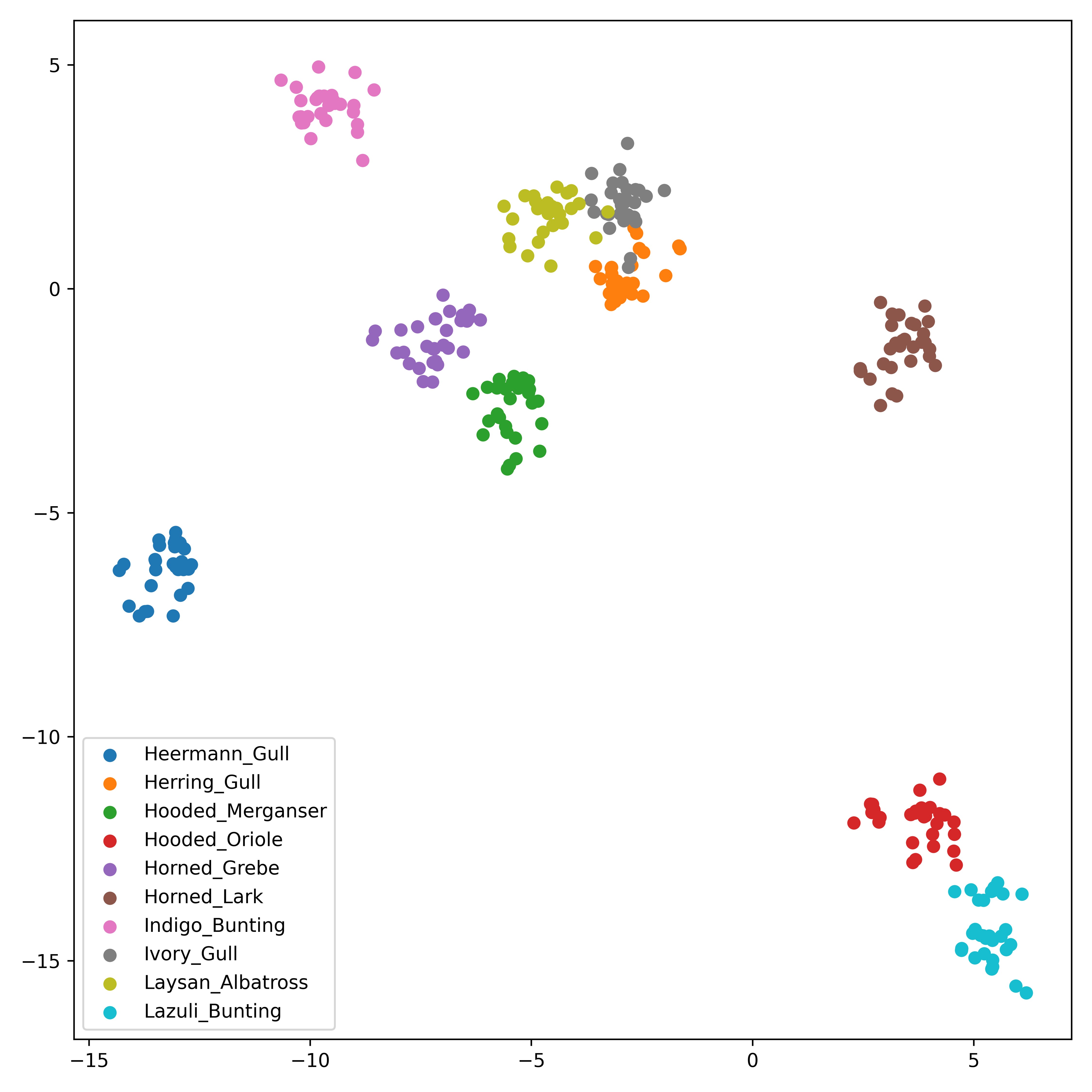}
    \caption{w/o $\mathcal{L}_{ED}$}
    \label{fig:EDloss_without}
    \end{subfigure}
    \hfill
    \begin{subfigure}[b]{0.49\columnwidth}
    \includegraphics[width=\columnwidth]{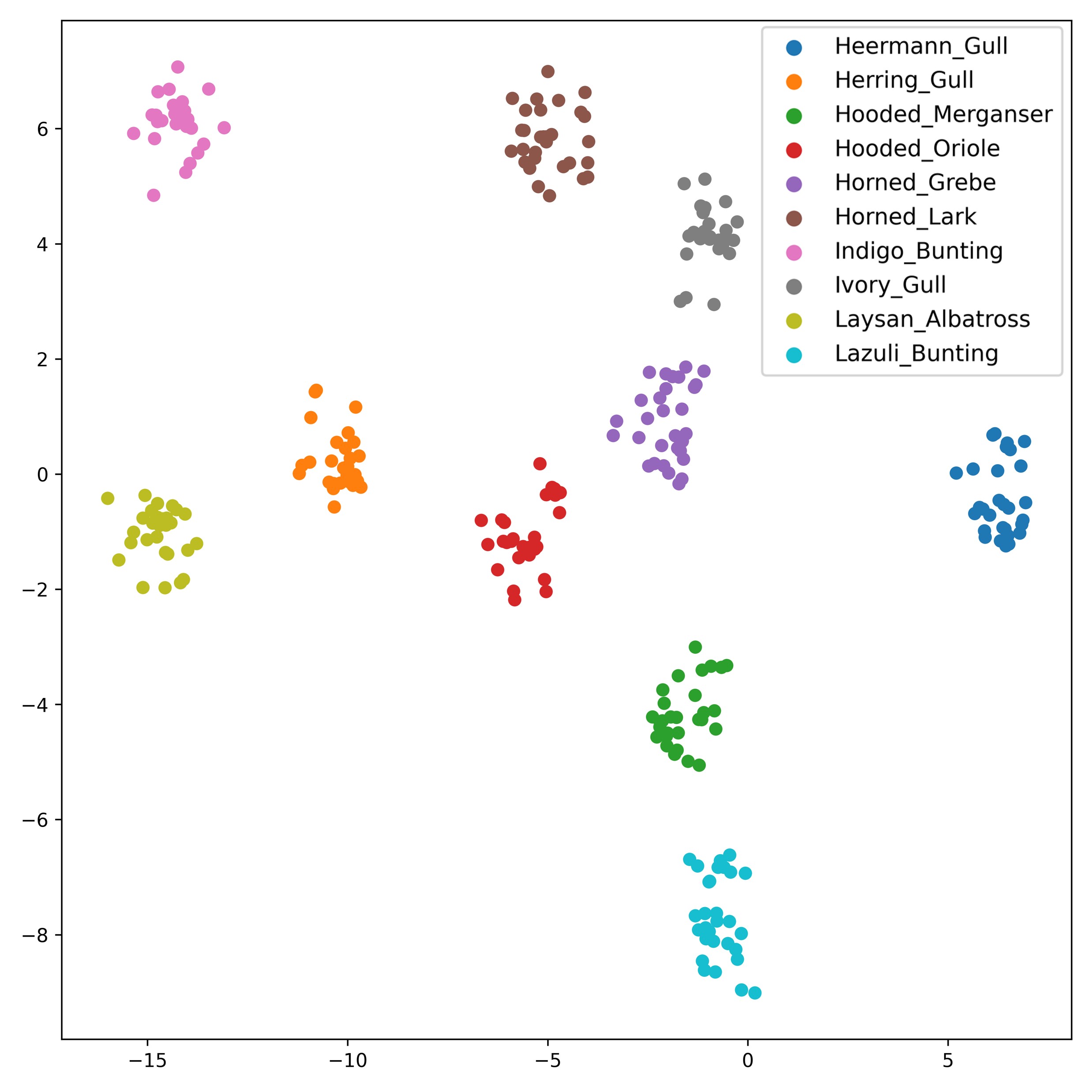}
    \caption{w/ $\mathcal{L}_{ED}$}
    \label{fig:EDloss_with}
    \end{subfigure}
    \vspace{-3mm}
    \caption{Feature space visualization on CUB200 to validate the efficacy of $\mathcal{L}_{ED}$. \vspace{-5mm}}
    \label{fig:EDloss}
\end{figure}
\vspace{-3mm}
\paragraph{Visualization to Validate Entropy-based Divergence Loss.}
To evaluate the effectiveness of the entropy-based divergence loss ($\mathcal{L}_{ED}$), we conducted additional experiments to visualize the feature space. The purpose of entropy-based divergence loss is to enhance the discriminative ability of the pre-trained ViT. We compared the feature space with and without the application of entropy-based divergence loss. Figure~\ref{fig:EDloss_without} displays the result without $\mathcal{L}_{ED}$, while Figure~\ref{fig:EDloss_with} shows the result with $\mathcal{L}_{ED}$.

As shown in Figure~\ref{fig:EDloss_without}, we observed certain classes with unclear decision boundaries, such as ``Herring Gull" and ``Ivory Gull". However, the application of the proposed entropy-based divergence loss clarifies the classification of these similar classes, as illustrated in Figure~\ref{fig:EDloss_with}. This result leads us to conclude that our proposed entropy-based divergence loss enhances the discriminative ability between similar classes. The improved discriminative power of the pre-trained ViT can contribute to the effective capture of domain-specific knowledge in the base session and it helps incremental session learning.

\vspace{-3mm}
\paragraph{Efficacy of Semantic Knowledge Distillation.}
We introduced semantic knowledge distillation loss ($\mathcal{L}_{SKD}$), utilizing language embeddings from a pre-trained language model to incorporate additional semantic knowledge. To evaluate the effectiveness of semantic knowledge distillation loss, we compared class-wise performance using a metric $\Delta_{diff}$, which represents the performance difference between results with and without $\mathcal{L}_{SKD}$.

\begin{figure}[t]
    \centering
    \includegraphics[width=0.95\columnwidth]{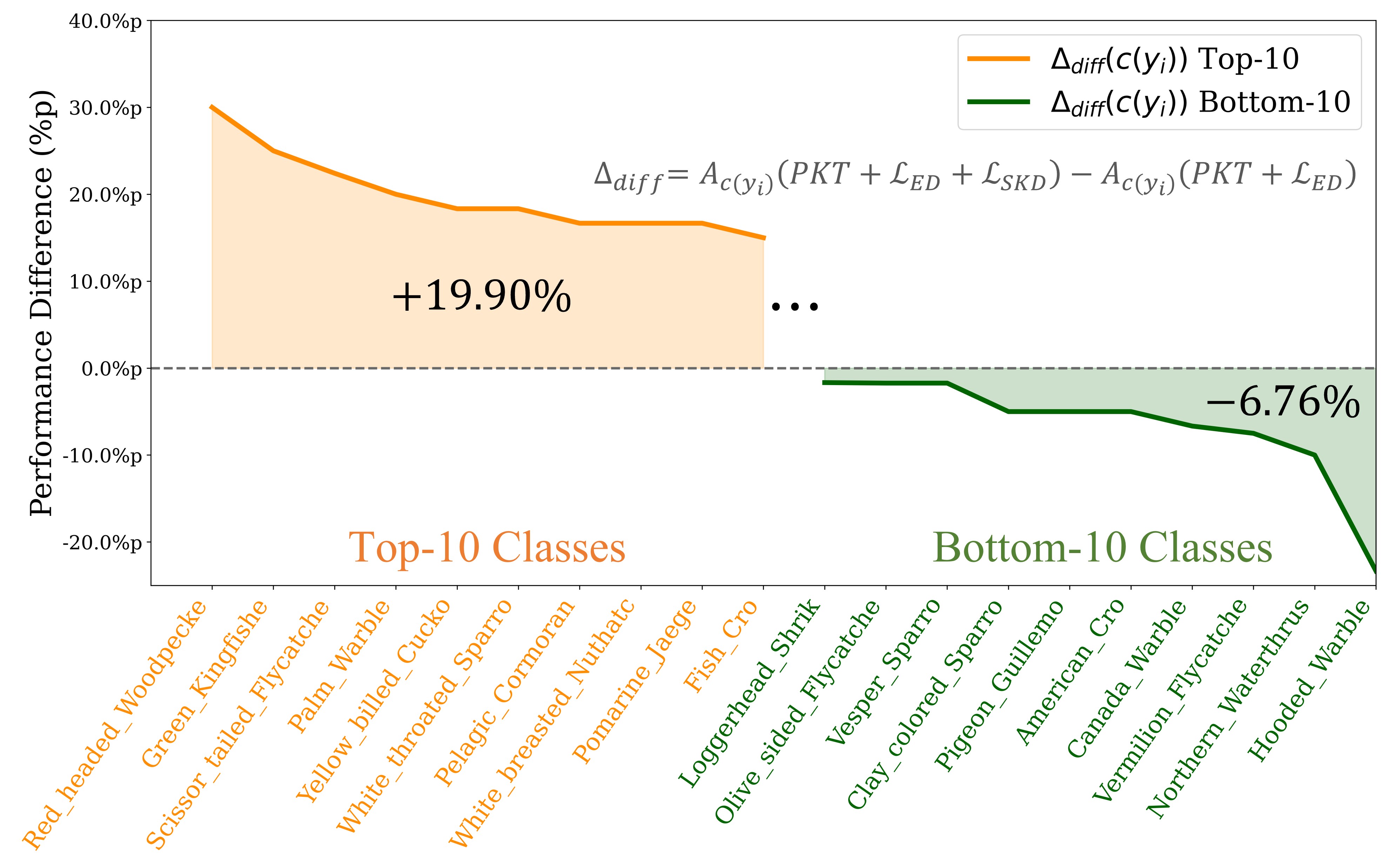}
    \vspace{-3mm}
    \caption{Class-wise accuracy on CUB200 to compare the performance of with and without $\mathcal{L}_{SKD}$.}
    \vspace{-6.mm}
    \label{fig:analysis-skd}
\end{figure}

In Figure~\ref{fig:analysis-skd}, we observed fluctuated performance with and without $\mathcal{L}_{SKD}$. For simplicity, we reported the top-10 classes with the most significant improvement and the bottom-10 classes with the most notable decline. As shown in Figure~\ref{fig:analysis-skd}, the top-10 classes exhibited an enhanced performance of about $19.90\%$. This improvement is noteworthy, even when considering that the bottom-10 classes showed a decrease of about $-6.76\%$. Particularly,  It is noteworthy that the names of the top-10 classes include color-related terms (\textit{e.g.}, red, green, yellow, and white) and shape descriptors (\textit{e.g.}, headed, tailed, palm, billed, pelagic, and fish). This observation underscores the high effectiveness of our $\mathcal{L}_{SKD}$, especially when the class names incorporate characteristic information. This effectiveness is especially valuable in the fine-grained dataset such as a CUB200.
\section{Conclusion}
\vspace{-2mm}
In this study, we introduced a novel Few-Shot Class Incremental Learning (FSCIL) method using large pre-trained vision and language transformers, coined PriViLege. We addressed severe issues of catastrophic forgetting and overfitting in Vision Transformer (ViT) through Pre-trained Knowledge Tuning (PKT), entropy-based divergence loss, and semantic knowledge distillation loss. Our proposed method, PriViLege, achieved significant performance improvements across all benchmarks, and we demonstrated our framework can be applicable to different pre-trained models, including ViT and CLIP. We believe our PriViLege sheds light on a new research direction utilizing large models in FSCIL research. As future work, we will explore how to apply pre-trained large models effectively to a challenging FSCIL scenario without the base session.
\vspace{-1mm}
\section*{Acknowledgements}
\vspace{-2mm}
This work was supported by MSIT (Ministry of Science and ICT), Korea, under the ITRC (Information Technology Research Center) support program (IITP-2024-RS-2023-00258649) supervised by the IITP (Institute for Information \& Communications Technology Planning \& Evaluation), and in part by the IITP grant funded by the Korea Government (MSIT) (Artificial Intelligence Innovation Hub) under Grant 2021-0-02068, and by the IITP grant funded by the Korea government (MSIT) (No.RS-2022-00155911, Artificial Intelligence Convergence Innovation Human Resources Development (Kyung Hee University)).
\newpage
{
    \small
    \bibliographystyle{ieeenat_fullname}
    \bibliography{main}

\begin{thebibliography}{42}
\providecommand{\natexlab}[1]{#1}
\providecommand{\url}[1]{\texttt{#1}}
\expandafter\ifx\csname urlstyle\endcsname\relax
  \providecommand{\doi}[1]{doi: #1}\else
  \providecommand{\doi}{doi: \begingroup \urlstyle{rm}\Url}\fi

\bibitem[Akata et~al.(2013)Akata, Perronnin, Harchaoui, and Schmid]{akata2013label}
Zeynep Akata, Florent Perronnin, Zaid Harchaoui, and Cordelia Schmid.
\newblock Label-embedding for attribute-based classification.
\newblock In \emph{Proceedings of the IEEE conference on computer vision and pattern recognition (CVPR)}, 2013.

\bibitem[Aky{\"u}rek et~al.(2022)Aky{\"u}rek, Aky{\"u}rek, Wijaya, and Andreas]{akyureksubspace}
Afra~Feyza Aky{\"u}rek, Ekin Aky{\"u}rek, Derry Wijaya, and Jacob Andreas.
\newblock Subspace regularizers for few-shot class incremental learning.
\newblock In \emph{International Conference on Learning Representations (ICLR)}, 2022.

\bibitem[Dosovitskiy et~al.(2021)Dosovitskiy, Beyer, Kolesnikov, Weissenborn, Zhai, Unterthiner, Dehghani, Minderer, Heigold, Gelly, Uszkoreit, and Houlsby]{dosovitskiy2021an}
Alexey Dosovitskiy, Lucas Beyer, Alexander Kolesnikov, Dirk Weissenborn, Xiaohua Zhai, Thomas Unterthiner, Mostafa Dehghani, Matthias Minderer, Georg Heigold, Sylvain Gelly, Jakob Uszkoreit, and Neil Houlsby.
\newblock An image is worth 16x16 words: Transformers for image recognition at scale.
\newblock In \emph{International Conference on Learning Representations (ICLR)}, 2021.

\bibitem[Gao et~al.(2023)Gao, Zhao, Sun, Xi, Zhang, Ghanem, and Zhang]{gao2023lae}
Qiankun Gao, Chen Zhao, Yifan Sun, Teng Xi, Gang Zhang, Bernard Ghanem, and Jian Zhang.
\newblock A unified continual learning framework with general parameter-efficient tuning.
\newblock \emph{International Conference on Computer Vision (ICCV)}, 2023.

\bibitem[He et~al.(2022)He, Zhou, Ma, Berg-Kirkpatrick, and Neubig]{he2022towards}
Junxian He, Chunting Zhou, Xuezhe Ma, Taylor Berg-Kirkpatrick, and Graham Neubig.
\newblock Towards a unified view of parameter-efficient transfer learning.
\newblock In \emph{International Conference on Learning Representations}, 2022.

\bibitem[Hersche et~al.(2022)Hersche, Karunaratne, Cherubini, Benini, Sebastian, and Rahimi]{hersche2022constrained}
Michael Hersche, Geethan Karunaratne, Giovanni Cherubini, Luca Benini, Abu Sebastian, and Abbas Rahimi.
\newblock Constrained few-shot class-incremental learning.
\newblock In \emph{Proceedings of the IEEE/CVF Conference on Computer Vision and Pattern Recognition}, 2022.

\bibitem[Hinton et~al.(2015)Hinton, Vinyals, and Dean]{hinton2015distilling}
Geoffrey Hinton, Oriol Vinyals, and Jeff Dean.
\newblock Distilling the knowledge in a neural network.
\newblock \emph{arXiv preprint arXiv:1503.02531}, 2015.

\bibitem[Huang et~al.(2024)]{huang2024learning}
Zitong Huang et~al.
\newblock Learning prompt with distribution-based feature replay for few-shot class-incremental learning.
\newblock \emph{arXiv preprint}, 2024.

\bibitem[Jung et~al.(2023)Jung, Han, Bang, and Song]{jung2023generating}
Dahuin Jung, Dongyoon Han, Jihwan Bang, and Hwanjun Song.
\newblock Generating instance-level prompts for rehearsal-free continual learning.
\newblock In \emph{Proceedings of the IEEE/CVF International Conference on Computer Vision}, 2023.

\bibitem[Kenton and Toutanova(2019)]{kenton2019bert}
Jacob Devlin Ming-Wei~Chang Kenton and Lee~Kristina Toutanova.
\newblock Bert: Pre-training of deep bidirectional transformers for language understanding.
\newblock In \emph{Proceedings of NAACL-HLT}, 2019.

\bibitem[Khan et~al.(2023)Khan, Naeem, Van~Gool, Stricker, Tombari, and Afzal]{khan2023introducing}
Muhammad Gul Zain~Ali Khan, Muhammad~Ferjad Naeem, Luc Van~Gool, Didier Stricker, Federico Tombari, and Muhammad~Zeshan Afzal.
\newblock Introducing language guidance in prompt-based continual learning.
\newblock In \emph{Proceedings of the IEEE/CVF International Conference on Computer Vision (ICCV)}, 2023.

\bibitem[Kim et~al.(2023)Kim, Han, Seo, and Moon]{kim2023warping}
Do-Yeon Kim, Dong-Jun Han, Jun Seo, and Jaekyun Moon.
\newblock Warping the space: Weight space rotation for class-incremental few-shot learning.
\newblock In \emph{The Eleventh International Conference on Learning Representations (ICLR)}, 2023.

\bibitem[Kingma and Ba(2014)]{kingma2014adam}
Diederik~P Kingma and Jimmy Ba.
\newblock Adam: A method for stochastic optimization.
\newblock \emph{arXiv preprint arXiv:1412.6980}, 2014.

\bibitem[Krizhevsky(2009)]{krizhevsky2009learning}
Alex Krizhevsky.
\newblock Learning multiple layers of features from tiny images.
\newblock 2009.

\bibitem[Kullback and Leibler(1951)]{kullback1951information}
Solomon Kullback and Richard~A Leibler.
\newblock On information and sufficiency.
\newblock \emph{The annals of mathematical statistics}, 1951.

\bibitem[Lester et~al.(2021)Lester, Al-Rfou, and Constant]{lester2021power}
Brian Lester, Rami Al-Rfou, and Noah Constant.
\newblock The power of scale for parameter-efficient prompt tuning.
\newblock In \emph{Proceedings of the 2021 Conference on Empirical Methods in Natural Language Processing (EMNLP)}, 2021.

\bibitem[Li and Liang(2021)]{li-liang-2021-prefix}
Xiang~Lisa Li and Percy Liang.
\newblock Prefix-tuning: Optimizing continuous prompts for generation.
\newblock In \emph{Proceedings of the 59th Annual Meeting of the Association for Computational Linguistics and the 11th International Joint Conference on Natural Language Processing (Volume 1: Long Papers)}, Online, 2021.

\bibitem[Loshchilov and Hutter(2017)]{loshchilovsgdr}
Ilya Loshchilov and Frank Hutter.
\newblock Sgdr: Stochastic gradient descent with warm restarts.
\newblock In \emph{International Conference on Learning Representations (ICLR)}, 2017.

\bibitem[Mazumder et~al.(2021)Mazumder, Singh, and Rai]{mazumder2021few}
Pratik Mazumder, Pravendra Singh, and Piyush Rai.
\newblock Few-shot lifelong learning.
\newblock In \emph{Proceedings of the AAAI Conference on Artificial Intelligence (AAAI)}, 2021.

\bibitem[McCloskey and Cohen(1989)]{mccloskey1989catastrophic}
Michael McCloskey and Neal~J Cohen.
\newblock Catastrophic interference in connectionist networks: The sequential learning problem.
\newblock In \emph{Psychology of learning and motivation}. 1989.

\bibitem[Moon et~al.(2023)Moon, Park, Kim, and Park]{moon2023online}
Jun-Yeong Moon, Keon-Hee Park, Jung~Uk Kim, and Gyeong-Moon Park.
\newblock Online class incremental learning on stochastic blurry task boundary via mask and visual prompt tuning.
\newblock In \emph{Proceedings of the IEEE/CVF International Conference on Computer Vision (ICCV)}, 2023.

\bibitem[Papyan et~al.(2020)Papyan, Han, and Donoho]{papyan2020prevalence}
Vardan Papyan, XY Han, and David~L Donoho.
\newblock Prevalence of neural collapse during the terminal phase of deep learning training.
\newblock \emph{Proceedings of the National Academy of Sciences}, 2020.

\bibitem[Peng et~al.(2022)Peng, Zhao, Wang, Li, and Lovell]{peng2022few}
Can Peng, Kun Zhao, Tianren Wang, Meng Li, and Brian~C Lovell.
\newblock Few-shot class-incremental learning from an open-set perspective.
\newblock In \emph{Computer Vision--ECCV 2022: 17th European Conference, Tel Aviv, Israel, October 23--27, 2022, Proceedings, Part XXV}, 2022.

\bibitem[Pennington et~al.(2014)Pennington, Socher, and Manning]{pennington2014glove}
Jeffrey Pennington, Richard Socher, and Christopher~D Manning.
\newblock Glove: Global vectors for word representation.
\newblock In \emph{Proceedings of the 2014 conference on empirical methods in natural language processing (EMNLP)}, 2014.

\bibitem[Pourpanah et~al.(2022)Pourpanah, Abdar, Luo, Zhou, Wang, Lim, Wang, and Wu]{pourpanah2022review}
Farhad Pourpanah, Moloud Abdar, Yuxuan Luo, Xinlei Zhou, Ran Wang, Chee~Peng Lim, Xi-Zhao Wang, and QM~Jonathan Wu.
\newblock A review of generalized zero-shot learning methods.
\newblock \emph{IEEE transactions on pattern analysis and machine intelligence (PAMI)}, 2022.

\bibitem[Radford et~al.(2021)Radford, Kim, Hallacy, Ramesh, Goh, Agarwal, Sastry, Askell, Mishkin, Clark, et~al.]{radford2021learning}
Alec Radford, Jong~Wook Kim, Chris Hallacy, Aditya Ramesh, Gabriel Goh, Sandhini Agarwal, Girish Sastry, Amanda Askell, Pamela Mishkin, Jack Clark, et~al.
\newblock Learning transferable visual models from natural language supervision.
\newblock In \emph{International Conference on Machine Learning (ICML)}, 2021.

\bibitem[Ravi and Larochelle(2017)]{ravi2017optimization}
Sachin Ravi and Hugo Larochelle.
\newblock Optimization as a model for few-shot learning.
\newblock In \emph{International conference on learning representations (ICLR)}, 2017.

\bibitem[Russakovsky et~al.(2015)Russakovsky, Deng, Su, Krause, Satheesh, Ma, Huang, Karpathy, Khosla, Bernstein, Berg, and Fei-Fei]{ILSVRC15}
Olga Russakovsky, Jia Deng, Hao Su, Jonathan Krause, Sanjeev Satheesh, Sean Ma, Zhiheng Huang, Andrej Karpathy, Aditya Khosla, Michael Bernstein, Alexander~C. Berg, and Li Fei-Fei.
\newblock {ImageNet Large Scale Visual Recognition Challenge}.
\newblock \emph{International Journal of Computer Vision (IJCV)}, 2015.

\bibitem[Seo et~al.(2023)Seo, Kang, and Park]{seo2023lfs}
Juwon Seo, Ji-Su Kang, and Gyeong-Moon Park.
\newblock Lfs-gan: Lifelong few-shot image generation.
\newblock In \emph{Proceedings of the IEEE/CVF International Conference on Computer Vision}, 2023.

\bibitem[Shi et~al.(2021)Shi, Chen, Zhang, Zhan, and Wu]{shi2021overcoming}
Guangyuan Shi, Jiaxin Chen, Wenlong Zhang, Li-Ming Zhan, and Xiao-Ming Wu.
\newblock Overcoming catastrophic forgetting in incremental few-shot learning by finding flat minima.
\newblock \emph{Advances in Neural Information Processing Systems (NeurIPS)}, 2021.

\bibitem[Smith et~al.(2023)Smith, Karlinsky, Gutta, Cascante-Bonilla, Kim, Arbelle, Panda, Feris, and Kira]{Smith_2023_CVPR}
James~Seale Smith, Leonid Karlinsky, Vyshnavi Gutta, Paola Cascante-Bonilla, Donghyun Kim, Assaf Arbelle, Rameswar Panda, Rogerio Feris, and Zsolt Kira.
\newblock Coda-prompt: Continual decomposed attention-based prompting for rehearsal-free continual learning.
\newblock In \emph{Proceedings of the IEEE/CVF Conference on Computer Vision and Pattern Recognition (CVPR)}, 2023.

\bibitem[Tang et~al.(2023)Tang, Peng, and Zheng]{tang2022learning}
Yu-Ming Tang, Yi-Xing Peng, and Wei-Shi Zheng.
\newblock When prompt-based incremental learning does not meet strong pretraining.
\newblock In \emph{Proceedings of the IEEE International Conference on Computer Vision (ICCV)}, 2023.

\bibitem[Tao et~al.(2020)Tao, Hong, Chang, Dong, Wei, and Gong]{tao2020few}
Xiaoyu Tao, Xiaopeng Hong, Xinyuan Chang, Songlin Dong, Xing Wei, and Yihong Gong.
\newblock Few-shot class-incremental learning.
\newblock In \emph{Proceedings of the IEEE/CVF Conference on Computer Vision and Pattern Recognition}, 2020.

\bibitem[Tian et~al.(2023)Tian, Li, Li, Ran, Ning, and Tiwari]{tian2023survey}
Songsong Tian, Lusi Li, Weijun Li, Hang Ran, Xin Ning, and Prayag Tiwari.
\newblock A survey on few-shot class-incremental learning.
\newblock \emph{arXiv preprint arXiv:2304.08130}, 2023.

\bibitem[Wah et~al.(2011)Wah, Branson, Welinder, Perona, and Belongie]{wah2011caltech}
Catherine Wah, Steve Branson, Peter Welinder, Pietro Perona, and Serge Belongie.
\newblock The caltech-ucsd birds-200-2011 dataset.
\newblock 2011.

\bibitem[Wang et~al.(2022{\natexlab{a}})Wang, Zhang, Ebrahimi, Sun, Zhang, Lee, Ren, Su, Perot, Dy, et~al.]{wang2022dualprompt}
Zifeng Wang, Zizhao Zhang, Sayna Ebrahimi, Ruoxi Sun, Han Zhang, Chen-Yu Lee, Xiaoqi Ren, Guolong Su, Vincent Perot, Jennifer Dy, et~al.
\newblock Dualprompt: Complementary prompting for rehearsal-free continual learning.
\newblock In \emph{European Conference on Computer Vision (ECCV)}, 2022{\natexlab{a}}.

\bibitem[Wang et~al.(2022{\natexlab{b}})Wang, Zhang, Lee, Zhang, Sun, Ren, Su, Perot, Dy, and Pfister]{wang2022learning}
Zifeng Wang, Zizhao Zhang, Chen-Yu Lee, Han Zhang, Ruoxi Sun, Xiaoqi Ren, Guolong Su, Vincent Perot, Jennifer Dy, and Tomas Pfister.
\newblock Learning to prompt for continual learning.
\newblock In \emph{2022 IEEE/CVF Conference on Computer Vision and Pattern Recognition (CVPR)}, 2022{\natexlab{b}}.

\bibitem[Yan et~al.(2021)Yan, Xie, and He]{yan2021dynamically}
Shipeng Yan, Jiangwei Xie, and Xuming He.
\newblock Der: Dynamically expandable representation for class incremental learning.
\newblock In \emph{Proceedings of the IEEE/CVF Conference on Computer Vision and Pattern Recognition (CVPR)}, 2021.

\bibitem[Yang et~al.(2022)Yang, Yuan, Li, Lin, Torr, and Tao]{yangneural}
Yibo Yang, Haobo Yuan, Xiangtai Li, Zhouchen Lin, Philip Torr, and Dacheng Tao.
\newblock Neural collapse inspired feature-classifier alignment for few-shot class-incremental learning.
\newblock In \emph{The Eleventh International Conference on Learning Representations}, 2022.

\bibitem[Zhang et~al.(2021)Zhang, Song, Lin, Zheng, Pan, and Xu]{zhang2021few}
Chi Zhang, Nan Song, Guosheng Lin, Yun Zheng, Pan Pan, and Yinghui Xu.
\newblock Few-shot incremental learning with continually evolved classifiers.
\newblock In \emph{Proceedings of the IEEE/CVF conference on computer vision and pattern recognition (CVPR)}, 2021.

\bibitem[Zhou et~al.(2022{\natexlab{a}})Zhou, Wang, Ye, Ma, Pu, and Zhan]{zhou2022forward}
Da-Wei Zhou, Fu-Yun Wang, Han-Jia Ye, Liang Ma, Shiliang Pu, and De-Chuan Zhan.
\newblock Forward compatible few-shot class-incremental learning.
\newblock In \emph{Proceedings of the IEEE/CVF Conference on Computer Vision and Pattern Recognition}, 2022{\natexlab{a}}.

\bibitem[Zhou et~al.(2022{\natexlab{b}})Zhou, Ye, Ma, Xie, Pu, and Zhan]{zhou2022few}
Da-Wei Zhou, Han-Jia Ye, Liang Ma, Di Xie, Shiliang Pu, and De-Chuan Zhan.
\newblock Few-shot class-incremental learning by sampling multi-phase tasks.
\newblock \emph{IEEE Transactions on Pattern Analysis and Machine Intelligence (PAMI)}, 2022{\natexlab{b}}.

\end{thebibliography}
}
\clearpage
\setcounter{page}{1}

\renewcommand{\thetable}{S\arabic{table}}
\renewcommand{\thefigure}{S\arabic{figure}}
\setcounter{figure}{0}
\setcounter{table}{0}

\setcounter{section}{0}

\maketitlesupplementary
\appendix

The supplementary materials cover baseline information~(\S\ref{sup:sec-A}), additional details on main experiments~(\S\ref{sup:sec-B}), and ablation studies~(\S\ref{sup:sec-C}) including pre-trained knowledge tuning~(\S\ref{sup:sec-C-1}), modulation prompts~(\S\ref{sup:sec-C-2}), semantic knowledge distillation loss~(\S\ref{sup:sec-C-3}), and further analysis for the performance (\S\ref{sup:sec-C-4}). Also included are PKT for domain-specific knowledge~(\S\ref{sup:sec-D}), analysis of transferability~(\S\ref{sup:sec-E}), limitation of prefix tuning~(\S\ref{sup:sec-F}), and visualization of entropy-based divergence loss effectiveness~(\S\ref{sup:sec-G}).

\section{Details of Baseline}
\label{sup:sec-A}
\paragraph{CEC.} In \citet{zhang2021few}, a graph attention network (GAN) acts as a classifier. The feature extractor is trained for strong base knowledge, and the GAN adapts to novel class knowledge in incremental sessions. GAN parameters increase with each new session. In our experiments, we partially trained the first two feature extractor layers for stable GAN training.
\vspace{-5mm}
\paragraph{NC-FSCIL.} \citet{yangneural} proposed the framework inspired by neural collapse (\citet{papyan2020prevalence}) which aims to align between the feature and corresponding weight of the classifier. NC-FSCIL pre-assigned the set of classifier prototypes which is formed as a simplex equiangular tight frame (ETF). NC-FSCIL proposed aligning a classifier with prototypes to enhance the performance of the classifier.
\vspace{-5mm}
\paragraph{WaRP.} \citet{kim2023warping} introduced the weight space rotation process which is called WaRP. They change the trained weight space into a new space where most of the important previous knowledge is condensed into a few parameters. It means WaRP can train the network to capture the knowledge in incremental sessions without suffering catastrophic forgetting.
\vspace{-5mm}
\paragraph{L2P.} 
In \citet{wang2022learning}, L2P is a prompt-based framework for class incremental learning, leveraging a pre-trained vision transformer. Using the prompt pool, L2P selects a prompt based on input samples and fine-tunes it for training. We adopted a single prompt during training to avoid performance deterioration from expanding prompts and omitted the prompt selection process.
\vspace{-5mm}
\paragraph{DualPrompt.} In \citet{wang2022dualprompt}, DualPrompt excels in class incremental learning by training G-Prompt and E-Prompt separately. It dynamically expands E-Prompt to retain task-specific knowledge. Similar to L2P, in our experiment, we used a single G-Prompt and E-Prompt, respectively and omitted prompt selection during evaluation.

\section{More Details for Main Experiments}
\label{sup:sec-B}
We presented the average accuracy across five simulations on CUB200, CIFAR-100, and miniImageNet. The highest and second-highest performances were indicated by bold and underlined text, respectively.

As shown in Table~\ref{tab:sup-cub200}. Our method outperformed others in all sessions, except the base session with ViT-B. In contrast to prior approaches suffering significant performance drops with new class arrivals, our method demonstrated a minor performance decline. Notably, prompt-based methods like L2P and DualPrompt performed less effectively than baselines. While prompt-based methods show promising performance in class incremental learning, their effectiveness diminished in FSCIL where transferability is crucial. Due to the limited trainable parameters of prompts, they struggle to capture sufficient domain-specific knowledge in the base session, impeding effective transfer to incremental sessions. 
Our method, PriViLege, designed to transfer diverse and domain-specific knowledge leveraging prompts, successfully mitigated catastrophic forgetting in FSCIL, aiding newly introduced classes.

In Table~\ref{tab:sup-cifar100}, we reported the performance on CIFAR-100. Our method, PriViLege, consistently outperformed other baselines in every incremental session. It is noteworthy that in training session 4, our method exhibited an enhanced performance of approximately $+0.28\%$ compared to the previous session. This improvement is particularly significant considering that WaRP, the second-highest performer, experienced a substantial performance decline of about $-2.11\%$ in the same session. The notable performance gain of our proposed method emphasizes its robust transferability, which not only contributes to forward transfer but also marginally contributes to positive backward transfer.

In Table~\ref{tab:sup-mini}, utilizing a network pre-trained on ImageNet-21K (\citet{ILSVRC15}), our method demonstrated the highest performance among all datasets. CEC exhibited a secondary performance, attributed to its partial network training. Surprisingly, our method reported minimal knowledge forgetting even after training all sessions. It is noteworthy that methods leveraging pre-trained knowledge, such as L2P and DualPrompt, showed competitive performance with existing FSCIL methods such as WaRP and NC-FSCIL. This observation underscores the significance of considering how to effectively leverage pre-trained knowledge when employing a ViT in FSCIL.

\begin{table*}[t]

\vspace{10mm}
\resizebox{\textwidth}{!}{%
\setlength{\tabcolsep}{2.pt}
\renewcommand{\arraystretch}{2.3}
\begin{tabular}{ccccccccccccc}
\specialrule{1.5pt}{1pt}{1pt}
                         & \multicolumn{11}{c}{\Large Sessions}  &  \\ \cline{2-12}
\multirow{-2}{*}{\Large Method} & {\large \text{$A_{Base}$}} & \large 1 & \large 2 & \large 3 & \large 4 & \large 5 & \large 6 & \large 7 & \large 8 & \large 9 & {\large \text{$A_{Last}$}} & \multirow{-2}{*}{\Large \text{{$A_{Avg}$}}} \\ \hline
Fine-Tuning + Proto $\psi$      &  \textbf{84.21±0.13} & 66.43±3.40 & 25.00±14.47 &  25.44±6.70 &  16.19±12.58 &  4.58±3.34  &  1.42±1.03  &  1.49±0.80  &  3.62±3.94  &  5.50±5.67  &  3.79±1.47  &  21.60±1.32 \\ \hdashline
CEC{[}CVPR'21{]}         &  75.40±8.01 &  \underline{73.23±8.32} &  \underline{72.00±8.25}  &  \underline{68.70±8.43} &  \underline{69.35±8.68}  &  \underline{67.78±7.88} &  \underline{67.01±7.79} &  \underline{66.40±8.04} &  \underline{65.78±8.10} &  \underline{65.57±7.95} &  \underline{65.70±8.03} &  \underline{72.41±1.18} \\
L2P{[}CVPR'22{]}         &  44.97±2.32 &  30.28±6.67 &  27.21±6.04  &  24.44±5.44 &  22.41±4.87  &  20.81±4.49 &  19.47±4.24 &  18.19±4.09 &  17.16±3.87 &  16.26±3.65 &  15.41±3.45 &  24.99±4.30 \\
DualPrompt[ECCV'22]   &  53.37±1.83 &  45.99±2.58 &  41.15±2.85   &  37.33±2.86 &  34.32±2.72   &  31.57±2.45 &  29.44±2.34 &  27.58±2.20 &  25.92±2.24 &  24.55±2.12 &  23.25±2.02 &  36.30±2.39 \\
NC-FSCIL{[}ICLR'23{]}    &  78.49±2.32 &  71.52±2.11 &  65.54±1.93   &  60.30±1.78 &  55.81±1.65   &  51.96±1.53 &  48.72±1.44 &  45.78±1.35 &  43.18±1.27 &  40.92±1.21 &  38.80±1.14 &  57.92±1.71 \\
WaRP{[}ICLR'23{]}        &  67.74±5.57 &  64.21±5.54 &  61.06±5.90   &  57.80±5.93 &  55.78±5.96   &  53.81±6.08 &  52.82±6.25 &  51.61±6.47 &  50.13±6.27 &  50.02±6.23 &  49.36±6.56 &  55.85±6.06 \\ \hdashline
\textbf{PriViLege (Ours)}          &  \underline{82.21±0.20} &  \textbf{81.25±0.20} & \textbf{80.45±0.20}   &  \textbf{77.76±0.41} &  \textbf{77.78±0.47}   &  \textbf{75.95±0.40} &  \textbf{75.69±0.41} &  \textbf{76.00±0.33} &  \textbf{75.19±0.45} &  \textbf{75.19±0.47} &  \textbf{75.08±0.52} &  \textbf{77.50±0.33} \\
\specialrule{1.5pt}{1pt}{1pt}
\end{tabular}
}
\caption{The performance of every session on CUB200.}
\label{tab:sup-cub200}
\bigskip
\resizebox{\textwidth}{!}{%
\setlength{\tabcolsep}{2.5pt}
\renewcommand{\arraystretch}{1.8}
\begin{tabular}{ccccccccccc}
\specialrule{1.5pt}{1pt}{1pt}
\multirow{2}{*}{\Large Method} & \multicolumn{9}{c}{\Large Sessions} &   \multirow{2}{*}{\Large \text{$A_{Avg}$}} \\ \cline{2-10}
                        & {\large \text{$A_{Base}$}}                & \large 1                   & \large 2                   & \large 3                   & \large 4                   & \large 5                   & \large 6                   & \large 7                   & {\large \text{$A_{Last}$}}                &                      \\ \hline
Fine-Tuning + Proto $\psi$     & \textbf{91.36±0.15} & 73.95±1.38          & 41.61±12.23         & 40.46±10.96         & 41.69±9.77          & 13.96±8.53          & 16.45±10.66         & 8.71±5.90           & 5.19±0.13           & 37.04±1.06           \\ \hdashline
CEC[CVPR'21]        & 74.20±2.03          & 71.49±2.13          & 70.11±2.54          & 67.34±2.88          & 65.96±2.64          & 65.14±3.36          & 64.74±3.96          & 63.48±4.09          & 61.48±3.33          & 67.10±2.92           \\
L2P[CVPR'22]        & 83.29±0.50          & 76.81±0.43          & 71.29±0.43          & 66.53±0.39          & 62.38±0.36          & 58.68±0.38          & 55.42±0.36          & 52.49±0.33          & 49.87±0.31          & 64.08±0.39           \\
DualPrompt[ECCV'22] & 85.11±0.29          & 78.42±0.29          & 72.81±0.35          & 67.92±0.35          & 63.69±0.29          & 59.92±0.26          & 56.60±0.23          & 53.62±0.21          & 50.93±0.21          & 65.45±0.27           \\
NC-FSCIL[ICLR'23]   & 89.51±0.23          & \underline{82.62±0.21}    & 76.72±0.19          & 71.61±0.18          & 67.13±0.17          & 63.18±0.16          & 59.67±0.15          & 56.53±0.14          & 53.70±0.14          & 68.96±0.17           \\
WaRP[ICLR'23]       & 86.20±1.46          & 82.58±1.53          & \underline{79.30±1.77}    & \underline{75.57±1.66}    & \underline{73.46±1.61}    & \underline{71.07±1.69}    & \underline{69.58±1.80}    & \underline{67.70±1.85}    & \underline{65.48±1.87}    & \underline{74.55±1.67}     \\ \hdashline
\textbf{PriViLege (Ours)}         & \underline{90.88±0.20}    & \textbf{89.39±0.23} & \textbf{88.97±0.15} & \textbf{87.55±0.24} & \textbf{87.83±0.24} & \textbf{87.35±0.24} & \textbf{87.53±0.25} & \textbf{87.15±0.21} & \textbf{86.06±0.32} & \textbf{88.08±0.20} \\
\specialrule{1.5pt}{1pt}{1pt}
\end{tabular}%
}
\caption{The performance of every session on CIFAR-100.}
\label{tab:sup-cifar100}
\bigskip
\resizebox{\textwidth}{!}{%
\setlength{\tabcolsep}{1.5pt}
\renewcommand{\arraystretch}{1.8}
\begin{tabular}{ccccccccccc}
\specialrule{1.5pt}{1pt}{1pt}
\multirow{2}{*}{\Large Method} & \multicolumn{9}{c}{\Large Sessions}                                                                                                                                                                              & \multirow{2}{*}{\Large \text{$A_{Avg}$}} \\ \cline{2-10}
                        & {\large \text{$A_{Base}$}}                & \large 1                   & \large 2                   & \large 3                   & \large 4                   & \large 5                   & \large 6                   & \large 7                   & {\large \text{$A_{Last}$}}                &                      \\ \hline
Fine-Tuning + Proto $\psi$    & 93.67±0.02       & 87.12±5.61          & 73.54±15.17         & 50.29±16.74         & 26.39±17.13         & 7.29±0.02           & 23.52±18.90         & 29.74±4.40          & 9.87±5.42           & 44.60±0.92           \\ \hdashline
CEC[CVPR'21]            & 87.43±5.90                & 85.99±6.70          & \underline{84.03±7.03}    & \underline{83.21±7.28}    & \underline{83.11±7.16}    & \underline{81.64±7.66}    & \underline{80.66±7.56}    & \underline{80.72±7.56}    & \underline{80.74±7.51}    & \underline{83.06±7.14}     \\
L2P[CVPR'22]            & 94.59±0.21                & 87.49±0.45          & 81.18±0.49          & 75.76±0.45          & 71.05±0.39          & 66.86±.0.36         & 63.15±0.34          & 59.82±0.32          & 56.84±0.32          & 72.97±0.36           \\
DualPrompt[ECCV'22]     & \underline{95.05±0.20}          & \underline{87.81±0.19}    & 81.51±0.21          & 76.07±0.21          & 71.38±0.12          & 67.19±0.15          & 63.45±0.12          & 60.15±0.10          & 57.14±0.11          & 73.31±0.15           \\
NC-FSCIL[ICLR'23]       & 77.25±0.42                & 71.30±0.39    & 66.21±0.36          & 61.80±0.34          & 57.94±0.32          & 54.53±0.30          & 51.50±0.28          & 48.79±0.27          & 46.35±0.25          & 59.52±0.33           \\
WaRP[ICLR'23]           & 83.30±1.06                & 80.53±1.48          & 77.22±1.01    & 74.99±1.50    & 73.64±0.97    & 71.52±1.07    & 69.16±0.84    & 68.79±0.79    & 67.97±1.28    & 74.13±1.08     \\ \hdashline
\textbf{PriViLege (Ours)}         & \textbf{96.68±0.06} & \textbf{96.49±0.05} & \textbf{95.65±0.15} & \textbf{95.54±0.13} & \textbf{95.54±0.13} & \textbf{94.91±0.16} & \textbf{94.33±0.15} & \textbf{94.19±0.12} & \textbf{94.10±0.13} & \textbf{95.27±0.11} \\
\specialrule{1.5pt}{1pt}{1pt}
\end{tabular}%
}
\caption{The performance of every session on miniImageNet.}
\label{tab:sup-mini}
\end{table*}

\begin{table}[t]
\resizebox{\columnwidth}{!}{%
\setlength{\tabcolsep}{2.pt}
\renewcommand{\arraystretch}{1.5}
\begin{tabular}{ccccccccc}
\specialrule{1.5pt}{1pt}{1pt}
\multicolumn{3}{c}{PKT Components} &  & \multicolumn{5}{c}{CUB200}                                            \\ \cline{1-3} \cline{5-9}
LT     & Modulation     & B+VL     &  & $A_{Base}$                &  & $A_{Last}$                &  & $A_{Avg}$                 \\ \cline{1-3} \cline{5-9}
       &                &          &  & \textbf{84.21±0.13} &  & 3.79±1.47           &  & 21.60±1.32          \\
       & \checkmark              &          &  & 65.31±1.81          &  & 51.04±1.36          &  & 57.47±1.51          \\
       &                & \checkmark        &  & 76.43±0.35          &  & 60.32±0.73          &  & 67.38±0.41          \\
       & \checkmark              & \checkmark        &  & 76.20±0.41          &  & 61.47±0.83          &  & 67.86±0.52          \\
\checkmark      &                &          &  & 74.48±0.14          &  & 64.75±0.99          &  & 68.66±0.52          \\
\checkmark      & \checkmark              &          &  & 77.38±0.82          &  & 68.09±1.02          &  & 71.42±0.80          \\
\checkmark      &                & \checkmark        &  & 78.30±1.55          &  & 68.58±2.68          &  & 72.07±1.90          \\ \hdashline
\checkmark      & \checkmark              & \checkmark        &  & 79.06±0.77          &  & \textbf{70.81±0.76} &  & \textbf{73.36±0.77} \\
\specialrule{1.5pt}{1pt}{1pt}
\end{tabular}%
}
\vspace{-3mm}
\caption{Further ablation experiment for PKT on CUB200. Modulation denotes leveraging modulation prompts and B+VL denotes prefix tuning the B-Prompt and prompt tuning the VL-Prompt.\vspace{-5mm}}
\label{tab:sup-pkt}
\end{table}

\section{Additional Ablation Studies}
\label{sup:sec-C}
\vspace{-1mm}
We conducted additional ablation studies to confirm the effectiveness of each proposed component, focusing on pre-trained knowledge tuning, modulation prompts, and semantic knowledge distillation loss.

\subsection{Ablation Study for PKT}
\label{sup:sec-C-1}
\vspace{-1mm}
In Table~\ref{tab:sup-pkt}, we conducted an ablation study on pre-trained knowledge tuning. Our baseline (row 3) utilized fine-tuning with a prototype classifier $\psi$. Rows 4 to 6 did not incorporate layer tuning. In Table~\ref{tab:sup-pkt}, we observed a gradual performance enhancement with the proposed pre-trained knowledge tuning. Notably, employing all proposed components showed the highest performance in both $A_{Last}$ and $A_{Avg}$.

We observed that the absence of layer tuning led to lower performance in the base session compared to its presence. This observation highlights the importance of layer tuning for acquiring sufficient domain-specific information, especially given the limited capacity of a fixed model. Additionally, relying solely on modulation prompts recorded lower performance than leveraging only learnable prompts like B-Prompt and VL-Prompt. Modulation prompts, designed to facilitate learnable prompt updates, struggled to provide useful knowledge for prefix tuning. Consequently, adopting only B-Prompt and VL-Prompt yielded better performance than solely relying on modulation prompts, emphasizing their additional capacity, irrespective of layer tuning.

Additionally, we observed that adopting modulation prompts with additional learnable prompts showed performance enhancement in $A_{Last}$ and $A_{Avg}$ regardless of layer tuning. Since the modulation prompts can contribute to the update of the learnable prompts, additional prompts, especially B-Prompt, can capture more effective domain-specific knowledge via the modulation prompts. Lastly, we observed the most promising performance when we adopted all the proposed components. The proposed pre-trained knowledge tuning can contribute to capturing effective domain-specific knowledge at the base session due to the additional B-Prompt and VL-Prompt assisted by the modulation prompts while preserving pre-trained knowledge through partial layer tuning.

\begin{table}[t]
\resizebox{\columnwidth}{!}{%
\setlength{\tabcolsep}{4.5pt}
\renewcommand{\arraystretch}{1.4}
\begin{tabular}{cccccccc}
\specialrule{1.5pt}{1pt}{1pt}
\multicolumn{2}{c}{Prefix Tuning} &  & \multicolumn{5}{c}{CUB200}                                            \\ \cline{1-2} \cline{4-8}
Key            & Value            &  & $A_{Base}$                &  & $A_{Last}$                &  & $A_{Avg}$                 \\ \cline{1-2} \cline{4-4} \cline{6-6} \cline{8-8}
\multicolumn{2}{c}{$P_{M}^G$}           &  & 78.29±0.27 &  & 69.14±0.63 &  & 72.34±0.30 \\
\multicolumn{2}{c}{$P_{M}^S$}           &  & 78.68±0.17          &  & 69.94±1.14          &  & 72.91±0.25          \\ \hdashline
$P_{M}^S$            & $P_{M}^G$              &  & \textbf{79.06±0.77} &  & \textbf{70.81±0.76} &  & \textbf{73.36±0.77} \\
\specialrule{1.5pt}{1pt}{1pt}
\end{tabular}%
}
\vspace{-3mm}
\caption{Further ablation experiment for the modulation prompts.}
\label{tab:sup-mod_prompt}
\vfill
\resizebox{\columnwidth}{!}{%
\setlength{\tabcolsep}{4.pt}
\renewcommand{\arraystretch}{1.2}
\begin{tabular}{cccccccc}
\specialrule{1.5pt}{1pt}{1pt}
\multicolumn{2}{c}{$\mathcal{L}_{SKD}$} &  & \multicolumn{5}{c}{CUB200}                 \\ \cline{1-2} \cline{4-8}
$\mathcal{L}_{KD}$         & $\mathcal{L}_{CE}$        &  & $A_{Base}$       &  & $A_{Last}$       &  & $A_{Avg}$        \\ \cline{1-2} \cline{4-4} \cline{6-6} \cline{8-8}
            &            &  & 79.06±0.77 &  & 70.81±0.76 &  & 73.36±0.77 \\ 
\checkmark           &            &  & 80.24±0.59 &  & 71.59±0.58 &  & 74.51±0.13 \\ \hdashline 
\checkmark           & \checkmark          &  & \textbf{82.10±0.57} &  & \textbf{73.44±0.40} &  & \textbf{76.27±0.30} \\
\specialrule{1.5pt}{1pt}{1pt}
\end{tabular}%
}
\vspace{-3mm}
\caption{Further ablation experiment for semantic knowledge distillation on CUB200.}
\vspace{-5mm}
\label{tab:sup-skd}
\end{table}

\subsection{Ablation Study for Modulation}
\label{sup:sec-C-2}
For an analysis of the modulation prompt, we conducted an additional ablation study on CUB200. We assessed the effectiveness of the head-specific prompt $P_{M}^S$ and the generic prompt $P_{M}^G$ by separately incorporating each prompt in prefix tuning. As shown in Table~\ref{tab:sup-mod_prompt}, relying solely on each head-specific prompt or generic prompt resulted in lower performance compared to leveraging both prompts simultaneously. Since the modulation prompts are constructed in different layers, with the head-specific prompt originating from the MSA layer and the generic prompt from the MLP layer, the head-specific prompt can contribute to scaling the attention score of the B-Prompt, capturing additional relationships between key vectors. Meanwhile, the generic prompt affords the incorporated knowledge with the B-Prompt through the scaling value vectors. Thus, we demonstrated that utilizing modulation prompts is highly beneficial to assist B-Prompt in prefix tuning.

\subsection{Ablation Study for \texorpdfstring{$\mathcal{L}_{SKD}$}{TEXT}}
\label{sup:sec-C-3}
The proposed semantic knowledge distillation loss comprises knowledge distillation loss and cross-entropy loss. To thoroughly assess its effectiveness, we conducted an additional ablation study using only PKT as the baseline. As indicated in Table~\ref{tab:sup-skd}, our proposed semantic knowledge distillation loss exhibited a gradual improvement in performance. Especially, it is noteworthy that the cross-entropy loss significantly contributes to performance enhancement across all metrics, due to the reduction of heterogeneity between two different spaces.

\begin{table}[t]

\vspace{-2.5mm}
\parbox{\columnwidth}{
\centering
\setlength{\tabcolsep}{8.pt}
\renewcommand{\arraystretch}{1.2}
\begin{tabular}{ccccc}
\specialrule{1.1pt}{1pt}{1pt}
CIFAR-100    & $A_{Base}$           & $A_{Last}$           & $A_{Avg}$   & $Fgt$         \\ \hline
CLIP-FT   & 79.43          & 33.64          & 50.65     &   45.79  \\
CLIP-LP   & 82.20          & 48.51          & 58.95     &   33.69  \\ 
LP-DiF*-CLIP   & 80.23          & 72.02          & 75.12   &  8.21     \\ 
\hdashline
\textbf{PriViLege-CLIP} & \textbf{84.25} & \textbf{78.35} & \textbf{77.16}  &  \textbf{5.90} \\
\specialrule{1.1pt}{1pt}{1pt}
\end{tabular}%
\vspace{-3mm}
\caption{Experiments of fine-tuning (FT) and linear proving (LP). The performance of LP-DiF* comes from the original paper.}
\label{tab:clip}
}
\vfill
\parbox{\columnwidth}{
\vspace{2mm}
\setlength{\tabcolsep}{12.pt}
\renewcommand{\arraystretch}{1.2}
\begin{tabular}{cccc}
\specialrule{1.1pt}{1pt}{1pt}
CUB200 & \multicolumn{1}{c}{$A_{Base}$} & \multicolumn{1}{c}{$A_{Last}$} & \multicolumn{1}{c}{$A_{Avg}$} \\ \hline
CEC (ViT-S)       & 78.51                    & 70.10                    & 72.93                   \\
WaRP (ViT-S)      & 72.56                    & 56.96                    & 62.54                   \\ \hdashline
\textbf{PriViLege (ViT-S)} & \textbf{80.25}           & \textbf{72.24}           & \textbf{74.89}          \\ \hline
CEC (ViT-L)       & 76.78                    & 69.45                    & 71.82                   \\
WaRP (ViT-L)      & 78.83                    & 62.64                    & 68.38                   \\ \hdashline
\textbf{PriViLege (ViT-L)} & \textbf{83.79}                    & \textbf{76.43}                    & \textbf{79.20}                \\
\specialrule{1.1pt}{1pt}{1pt}
\end{tabular}%
\vspace{-3mm}
\caption{Experiments of adopting ViT-S and ViT-L on CUB200.\vspace{-3mm}}
\label{tab:vit_var}
}
\end{table}

\subsection{Further Analysis for the Performance}
\label{sup:sec-C-4}
We conducted further experiment to compare our method with the LP-DiF~\cite{huang2024learning} which is based on CLIP. Table~\ref{tab:clip} shows that PriViLege recorded better performance and lower forgetting than LP-DiF. We also analyzed the scalability of our proposed method. In Table~\ref{tab:vit_var}, our method showed improved performance when the capacity of the base model is increased. Experimental results shows that our method can expect performance enhancement in a more strong base model.

\subsection{Considering Pre-trained Base Model}
\label{sup:sec-C-5}
The pre-trained dataset, ImageNet-21K, includes almost all classes in the dataset used in experiments. For example, CIFAR-100 and CUB200 include 12 and 150 exclusive classes, respectively. We conducted further experiments to prove that the performance enhancement stems from the proposed method, PriViLege. Table~\ref{tab:wiki} shows our superior performance on FGVC-aircraft, a dataset non-overlapped with ImageNet-21K. Moreover, Table~\ref{tab:scratch} presents our remarkable performance even when training from scratch. These results support the superiority of our method regardless of the base model.
\begin{table}[t]
\vspace{-2.4mm}
\parbox{\columnwidth}{
\centering
\setlength{\tabcolsep}{11.5pt}
\renewcommand{\arraystretch}{1.2}
\begin{tabular}{cccc}
\specialrule{1.1pt}{1pt}{1pt}
FGVC-aircraft           & $A_{Base}$   & $A_{Last}$  & $A_{Avg}$  \\ \hline
CEC              &    23.05  &   16.85       &   19.46     \\
WaRP             &    24.85 &   15.69      &   19.67     \\ \hdashline
\textbf{PriViLege (Ours)} &    \textbf{58.30} &   \textbf{45.55}      &   \textbf{50.87}    \\
\specialrule{1.1pt}{1pt}{1pt}
\end{tabular}
\vspace{-4mm}
\caption{Experiments on FGVC-aircraft.}
\vspace{2mm}
\label{tab:wiki}
}
\vfill
\parbox{\columnwidth}{
\centering
\setlength{\tabcolsep}{15.pt}
\renewcommand{\arraystretch}{1.2}
\begin{tabular}{cccc}
\specialrule{1.1pt}{1pt}{1pt}
CIFAR-100 & \multicolumn{1}{l}{$A_{Base}$} & \multicolumn{1}{l}{$A_{Last}$} & \multicolumn{1}{l}{$A_{Avg}$} \\ \hline
CEC   &  8.30  & 4.76   & 6.09 \\ 
WaRP  &  35.82 & 23.30  & 28.85 \\ \hdashline
\textbf{PriViLege} & \textbf{50.37}   & \textbf{30.83}  & \textbf{39.05}                   \\
\specialrule{1.1pt}{1pt}{1pt}
\end{tabular}
\vspace{-5mm}
\caption{Experiments of ViT-B scratch.\vspace{-5mm}}
\label{tab:scratch}
}
\end{table}

\begin{figure}[t]
    \centering
    \includegraphics[width=0.8\columnwidth]{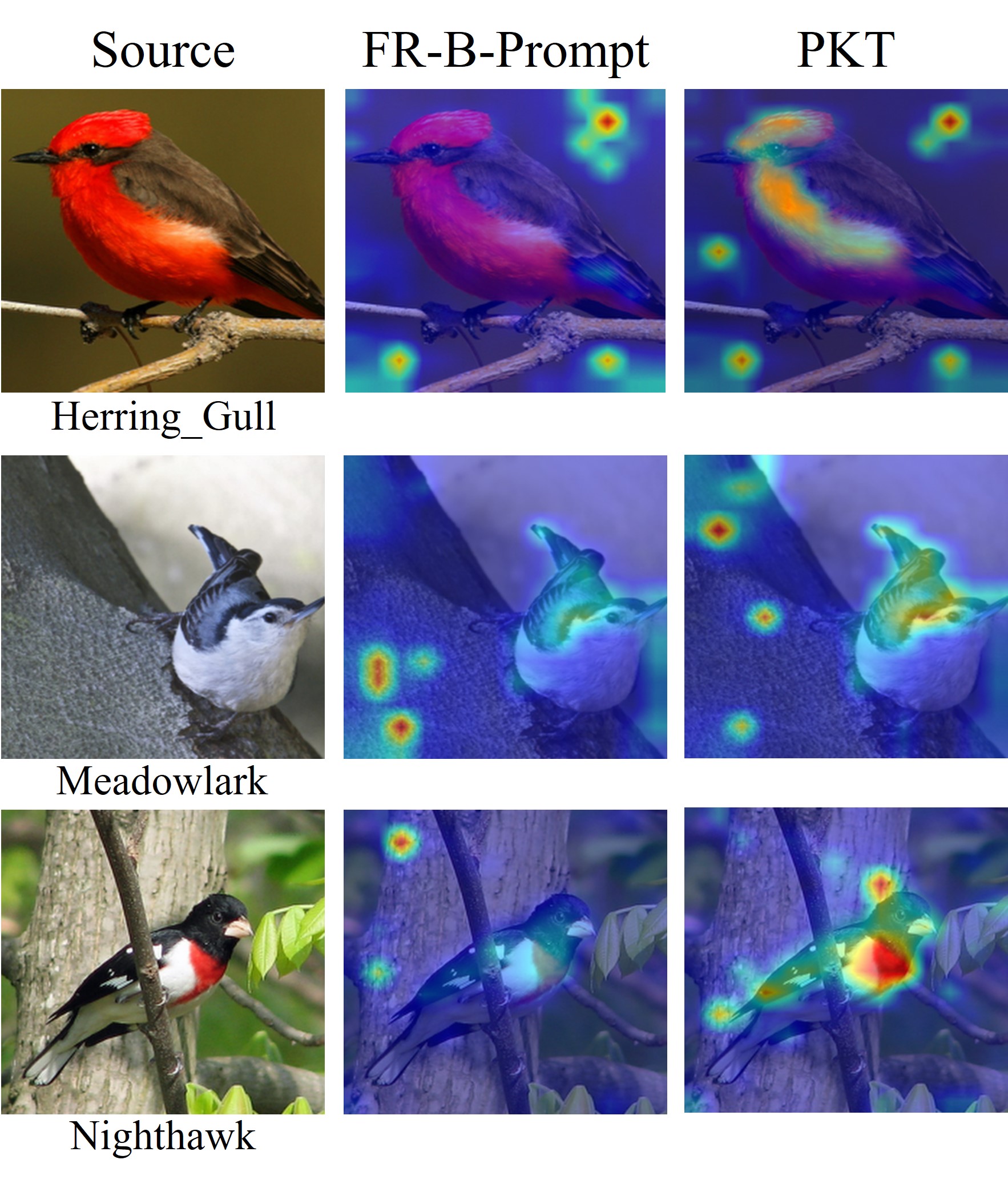}
    \vspace{-4mm}
    \caption{Attention map to assess the effectiveness of the proposed PKT on CUB200. FR-B-Prompt denotes prefix tuning learnable prompt with frozen ViT.}
    \label{fig:sup-attn}
    \vspace{-6mm}
\end{figure}

\vspace{-2mm}
\section{PKT for Domain-Specific Knowledge}
\label{sup:sec-D}
To confirm the effectiveness of pre-trained knowledge tuning in capturing domain-specific knowledge, we conducted additional analysis on CUB200. This aimed to clarify the reasons for performance enhancement through the proposed PKT. Figure~\ref{fig:sup-attn} displays the attention map of FR-B-Prompt, representing the use of a learnable prompt via prefix tuning on the frozen ViT, alongside the map of our proposed PKT. While FR-B-Prompt trained learnable parameters like B-Prompt using prefix tuning through the frozen ViT, the proposed PKT trained B-Prompt and modulation prompt with prefix tuning and also fine-tuned partial layers. For visualization, we trained a learnable B-Prompt but constructed the attention mask using only image tokens.
\begin{figure*}[t]
    \begin{subfigure}[b]{0.48\textwidth}
    \centering
    \includegraphics[width=\columnwidth]{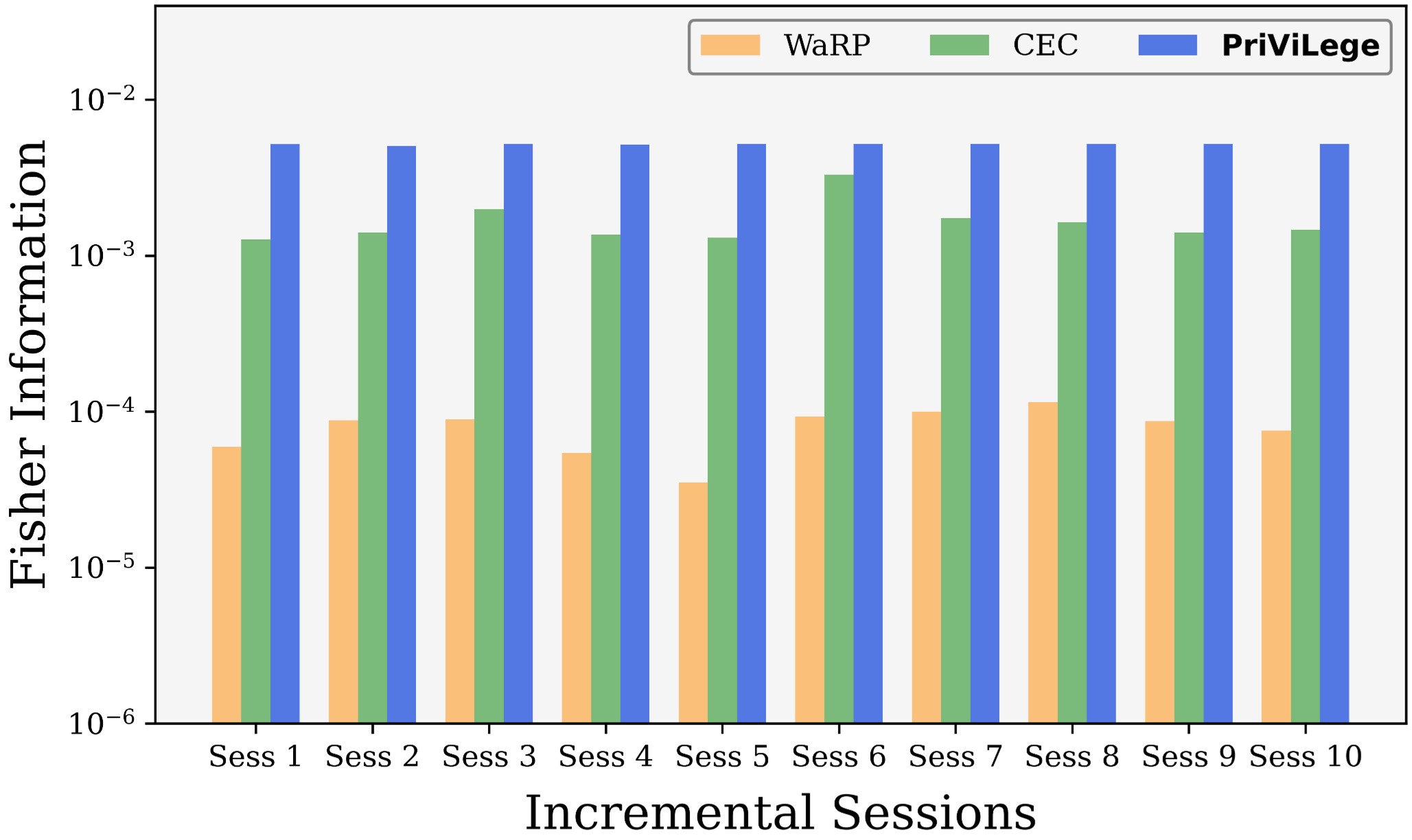}
    \caption{Comparison of the fisher information to assess the transferability.}
    \label{fig:sup-fisher}
    \end{subfigure}
    \hfill
    \begin{subfigure}[b]{0.48\textwidth}
    \centering
    \includegraphics[width=\columnwidth]{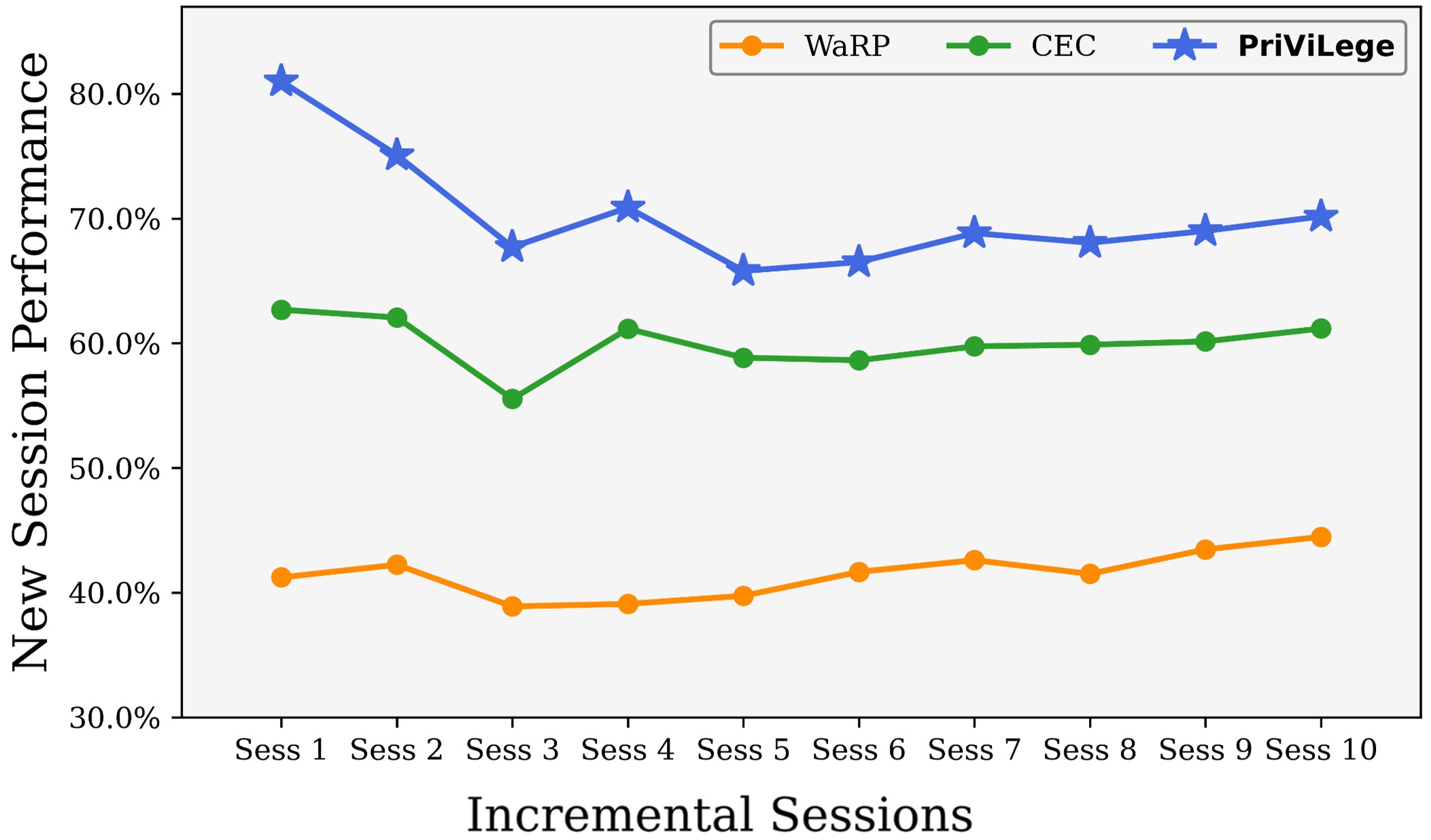}
    \caption{Comparison of the new task performance on CUB200.}
    \label{fig:sup-new_perform}
    \end{subfigure}
\caption{Comparison of the fisher information and new task performance on CUB200.}
\label{fig:sup-transfer}
\end{figure*}
As illustrated in Figure~\ref{fig:sup-attn}, we observed that the attention map of our proposed PKT exhibited greater activation towards the object compared to FR-B-Prompt. Unlike the attention map of FR-B-Prompt, the attention map of our proposed PKT was more focused on the object rather than the background. Through this observation, we demonstrate that our proposed PKT primarily aims to extract knowledge from the object. Since our PKT fine-tuned some pre-trained layers and trained B-prompt using the modulation prompt, which facilitated prefix tuning, our PKT can effectively capture domain-specific knowledge through the fine-tuned layers and more efficiently through the B-Prompt facilitated by the modulation prompts. Thus, our PKT can capture more domain-specific knowledge focused on the class object.

\section{Further Analysis for Transferability}
\label{sup:sec-E}
\vspace{-2mm}
 To compare the transferability of our method with state-of-the-art FSCIL methods, we conducted additional analyses, considering Fisher information (Figure~\ref{fig:sup-fisher}) and new task performance (Figure~\ref{fig:sup-new_perform}). Fisher information is widely used in continual learning as a metric to estimate how important the trained parameters are for the training of a given task. We assessed transferability through the Fisher information of the parameters trained at the base session. If the parameters trained at the base session have high value of the Fisher information in the incremental session, it indicates their importance for the incremental session. Through this, we evaluated the transferability of our method compared to other baselines. Additionally, by analyzing new task performance, we demonstrate effective incremental session learning through the transferred knowledge.

As illustrated in Figure~\ref{fig:sup-fisher}, our method, PriViLege, achieved the highest value of Fisher information compared to other baselines. This observation indicates that our method can effectively transfer useful domain-specific knowledge to incremental sessions. We demonstrated that our method captures transferable knowledge at the base session, consistently utilized as valuable knowledge for the incremental sessions. Furthermore, as shown in Figure~\ref{fig:sup-new_perform}, our method also reported the most promising new task performance. Given that new task performance measures the accuracy of each session under all seen classes, PriViLege demonstrated remarkable performance despite the few-shot data given at the incremental sessions. Through this observation, we also validated that our proposed method, which captures transferable and useful domain-specific knowledge, exhibits outstanding transferability to facilitate the learning of incremental sessions.

\begin{figure}[t]
\vspace{-2mm}
    \centering
    \includegraphics[width=1.\columnwidth]{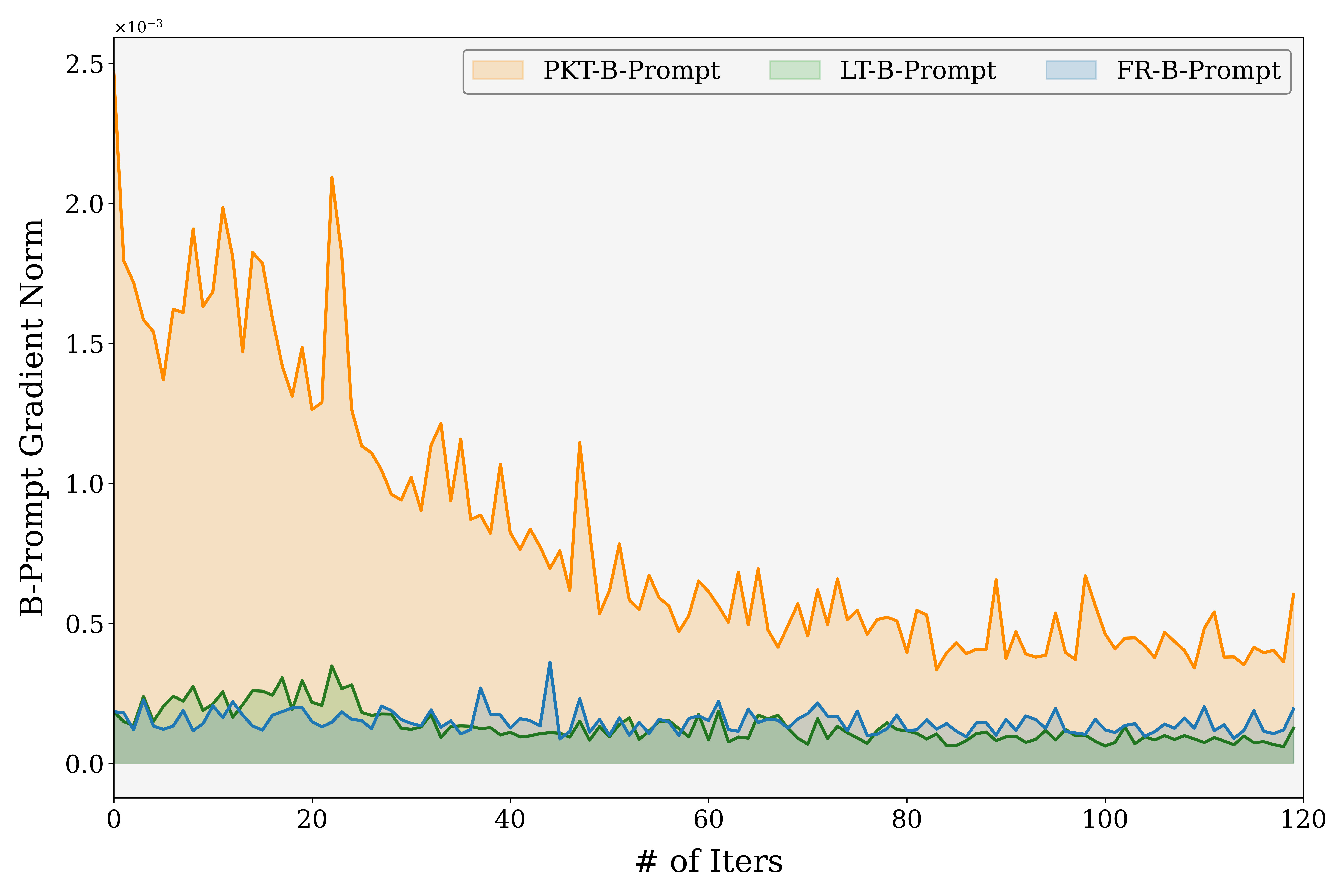}
    \vspace{-7mm}
    \caption{
B-Prompt gradient magnitudes on CUB200. PKT-B-Prompt, LT-B-Prompt, and FR-B-Prompt denote adopting PKT, only layer tuning, and fixed ViT for the B-Prompt.}
    \label{fig:sup-grad}
    \vspace{-3mm}
\end{figure}

\section{Analysis for the Limitation of Prefix Tuning}
\label{sup:sec-F}
As mentioned in Section~\ref{sec:PKT}, prefix tuning has a limitation in updating B-Prompt due to its slow adaptation speed. To overcome this limitation, we proposed modulation prompts. We conducted further analysis to validate that our proposed modulation prompts can effectively enhance the update of the B-Prompt. We calculated the norm of gradient vectors of B-Prompt at every iteration.

As illustrated in Figure~\ref{fig:sup-grad}, relying solely on layer tuning with B-Prompt or leveraging the frozen ViT showed a small norm of gradient vectors due to the slow adaptation speed of prefix tuning. This is because the feature vectors from the B-Prompt tokens are overwhelmed by feature vectors from the input tokens, causing the B-Prompt to struggle to contribute to capturing knowledge and suffer slow adaptation via prefix tuning. However, utilizing PKT, including the modulation prompts, demonstrated a promising increase in the gradient norm of B-Prompt. Since the modulation prompts can scale the key and value of the B-Prompt, it promotes the update of the B-Prompt effectively.

\begin{figure}[t]
    \begin{subfigure}[b]{0.48\columnwidth}
    \centering
    \includegraphics[width=1.\columnwidth]{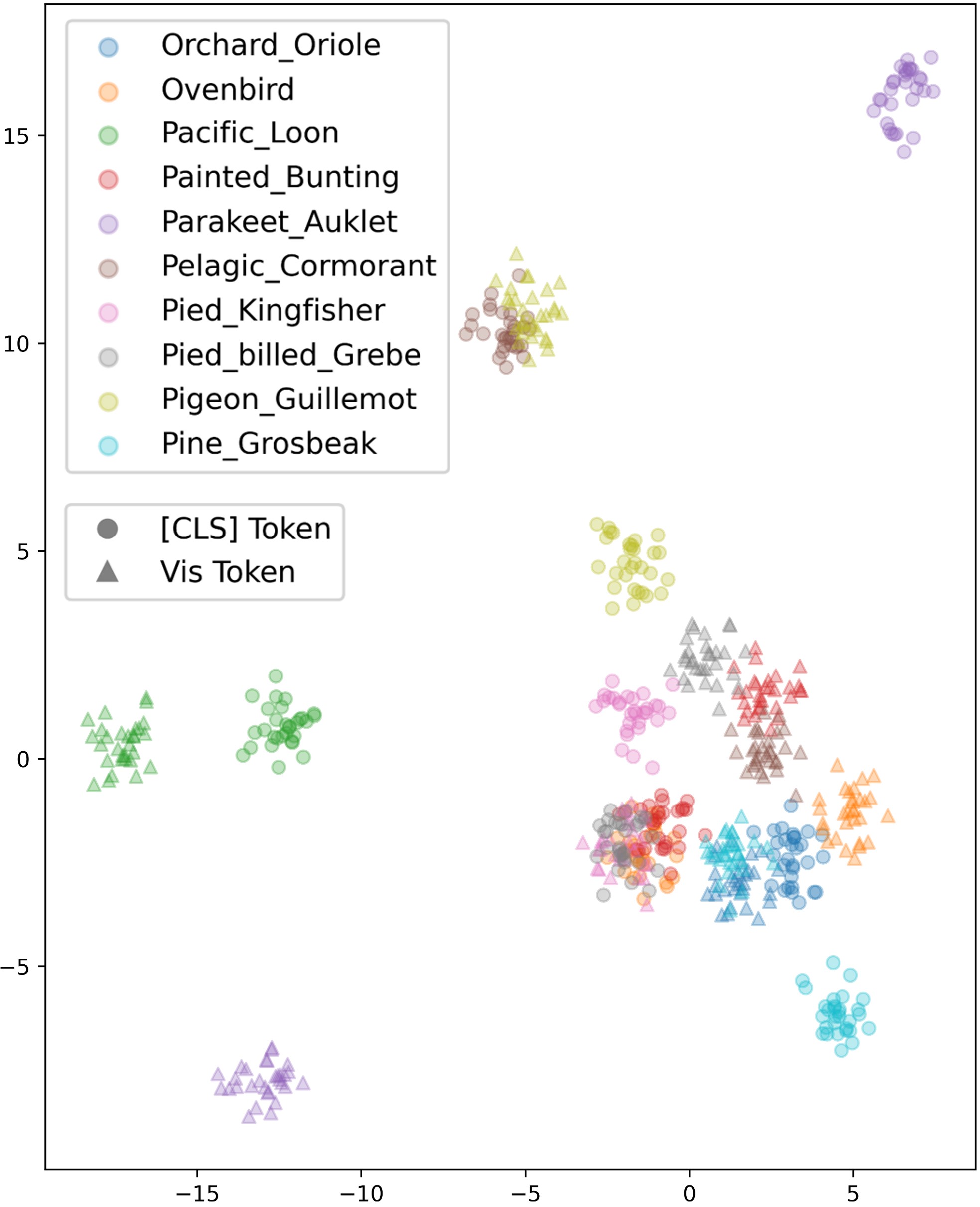}
    \caption{Without $\mathcal{L}_{ED}$.}
    \label{fig:sup-without_ed}
    \end{subfigure}
    \hfill
    \begin{subfigure}[b]{0.48\columnwidth}
    \centering
    \includegraphics[width=1.\columnwidth]{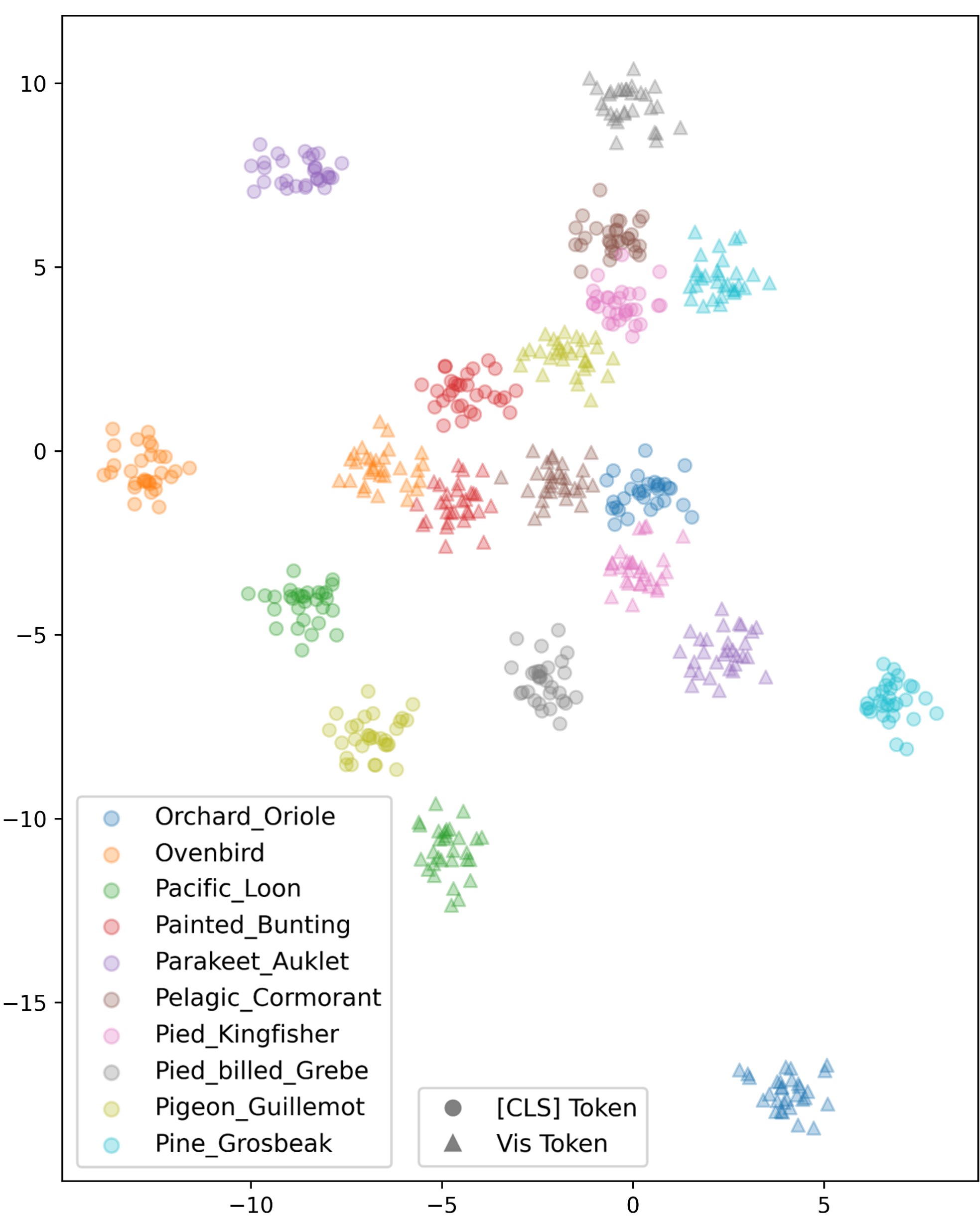}
    \caption{With $\mathcal{L}_{ED}$.}
    \label{fig:sup-with_ed}
    \end{subfigure}
    \vspace{-1.8mm}
    \caption{Feature space visualization of [CLS] and vision token on CUB200. The circle and triangle denote the [CLS] token and the vision token, respectively. Each color represents the classes.}
    \vspace{-3mm}
    \label{fig:sup-tsne}
\end{figure}

\section{Further Analysis the Effectiveness of \texorpdfstring{$\mathcal{L}_{ED}$}{TEXT}}
\label{sup:sec-G}
As mentioned in Section~\ref{sec:ED}, the average pooling of [CLS] and vision token results in sharing similar feature knowledge between [CLS] and vision token, hindering effective learning of the vision token. We further analyzed the problem of average pooling and the effectiveness of entropy-based divergence loss ($\mathcal{L}_{ED}$) in the perspective of feature vectors from [CLS] and vision token, respectively. As shown in Figure~\ref{fig:sup-tsne}, we visualized the feature space that includes [CLS] and vision token. Figure~\ref{fig:sup-without_ed} showed the feature space without applying entropy-based divergence loss, and Figure~\ref{fig:sup-with_ed} illustrated the feature space applying entropy-based divergence loss.

As shown in Figure~\ref{fig:sup-without_ed}, [CLS] and vision token are located closely in the feature space or even overlap with other classes. Since [CLS] and vision token share the same objective for the classification task due to average pooling, they struggle to capture discriminative knowledge to distinguish each other. This problem hinders the vision token from capturing effective knowledge and learning discriminative features for the classification task.

However, as illustrated in Figure~\ref{fig:sup-with_ed}, applying entropy-based divergence loss can effectively mitigate the problem of proximity or overlap. It is noteworthy that entropy-based divergence loss can also help [CLS] feature vector and vision feature vector become discriminative not only when they belong to different classes but also when they belong to the same class. Through this observation, we demonstrated the problem of average pooling and validated the effectiveness of the proposed entropy-based divergence loss. Our entropy-based divergence loss helps mitigate the sharing of knowledge between [CLS] and vision token and enhances discriminative ability, even when classifying [CLS] and vision token that belong to the same class.

\end{document}